\documentclass[conference,compsoc]{IEEEtran}
\ifCLASSOPTIONcompsoc
  \usepackage[nocompress]{cite}
\else
  \usepackage{cite}

\fi
\ifCLASSINFOpdf
\else
\fi

\usepackage{textcomp}
\usepackage[compact]{titlesec} 

\usepackage{booktabs}
\usepackage{multirow}
\usepackage{subcaption}
\usepackage{xcolor}
\usepackage{soul}
\usepackage{listings}
\usepackage{algorithm}
\usepackage{algorithmicx,algcompatible}
\usepackage{tikz}
\usetikzlibrary{positioning,arrows.meta}
\usepackage{hyperref}
\usepackage{enumitem}
\usepackage{amsmath}
\usepackage{amssymb}
\usepackage[capitalise, nameinlink]{cleveref}
\usepackage{algpseudocode}
\usepackage[table]{xcolor}

\usepackage{listings}
\usepackage{xcolor}
\usepackage{xspace}
\usepackage{minted}
\usepackage{xcolor}
\usepackage{booktabs}
\usepackage{multirow}
\usepackage{pifont}

\makeatletter
\crefname{algocfline}{line}{lines}
\Crefname{algocfline}{Line}{Lines}
\makeatother
\crefname{algorithm}{Algorithm}{Algorithms}
\Crefname{algorithm}{Algorithm}{Algorithms}
\crefname{section}{\S\!}{\S\S\!}
\Crefname{section}{Section}{Sections}
\crefname{figure}{Figure}{Figures}
\Crefname{figure}{Figure}{Figures}
\crefname{equation}{Equation}{Equations}
\Crefname{equation}{Equation}{Equations}
\crefname{listing}{Listing}{Listings}
\Crefname{listing}{Listing}{Listings}
\crefname{defn}{definition}{definitions}
\crefname{AlgoLine}{line}{lines}
\Crefname{AlgoLine}{Line}{Lines}
\crefname{algocfline}{line}{lines}
\Crefname{algocfline}{Line}{Lines}

\crefalias{algocfline}{line}

\definecolor{bg}{rgb}{0.97,0.97,0.97}

\setminted{
    fontsize=\footnotesize,
    bgcolor=bg,
    frame=lines,
    framesep=2mm,
    baselinestretch=1.1,
    breaklines=true,
    linenos,
    numbersep=5pt
}

\usepackage[T1]{fontenc}
\usepackage{textcomp}
\usepackage{xcolor}
\usepackage{listings}
\usepackage[most]{tcolorbox}

\definecolor{codeframe}{RGB}{210,210,210}
\definecolor{origbg}{RGB}{235,248,241}
\definecolor{engbg}{RGB}{255,247,230}
\definecolor{keywordblue}{RGB}{35,95,160}
\definecolor{commentgray}{RGB}{95,95,95}
\definecolor{stringorange}{RGB}{180,105,20}

\lstdefinestyle{pycompact}{
  language=Python,
  basicstyle=\ttfamily\scriptsize,
  keywordstyle=\color{keywordblue}\bfseries,
  commentstyle=\color{commentgray}\itshape,
  stringstyle=\color{stringorange},
  numbers=left,
  numberstyle=\tiny\color{gray},
  numbersep=6pt,
  frame=none,
  breaklines=true,
  showstringspaces=false,
  tabsize=2,
  columns=fullflexible,
  keepspaces=true,
  upquote=true
}

\newtcblisting{codepanel}[2]{
  colback=#1,
  colframe=codeframe,
  boxrule=0.4pt,
  arc=1.5pt,
  left=2pt,
  right=2pt,
  top=2pt,
  bottom=2pt,
  listing only,
  listing options={style=pycompact},
  title={#2},
  fonttitle=\bfseries\footnotesize,
  coltitle=black,
  enhanced,
  attach boxed title to top left={xshift=2mm,yshift=-1mm},
  boxed title style={
    colback=white,
    colframe=codeframe,
    boxrule=0.3pt,
    arc=1pt
  }
}

\newcommand{\para}[1]{\smallskip\noindent\textbf{#1}}

\newcommand{\sysname}{{\textsc{AIChilles}}\xspace}

\newcommand{\adrs}{{\textsc{ADSO}}\xspace}

\definecolor{promptaccent}{RGB}{63, 81, 181}       
\definecolor{promptbg}{RGB}{250, 250, 252}         
\definecolor{prompttitlebg}{RGB}{63, 81, 181}      
\definecolor{keywordcolor}{RGB}{183, 28, 28}       
\definecolor{codeexcolor}{RGB}{27, 94, 32}         
\definecolor{importantcolor}{RGB}{191, 54, 12}     
\definecolor{labelcolor}{RGB}{69, 90, 100}         

\newcommand{\pkey}[1]{\textcolor{keywordcolor}{\textbf{#1}}}
\newcommand{\pcode}[1]{{\small\textcolor{codeexcolor}{\texttt{#1}}}}
\newcommand{\pimp}[1]{\textcolor{importantcolor}{\textbf{#1}}}
\newcommand{\plabel}[1]{\textcolor{labelcolor}{\textit{#1}}}

\newtcolorbox[auto counter]{PromptBlock}[2][]{
    enhanced,
    breakable,
    colback=promptbg,
    colframe=promptaccent,
    coltitle=white,
    fonttitle=\bfseries\sffamily,
    title={Prompt~\thetcbcounter: #2},
    label={#1},
    boxrule=0.5pt,
    arc=1pt,
    left=10pt,
    right=10pt,
    top=8pt,
    bottom=8pt,
    toptitle=4pt,
    bottomtitle=4pt,
    colbacktitle=prompttitlebg,
    sharp corners=downhill,
    borderline west={3pt}{0pt}{promptaccent}
}

\tikzset{
    boxframe/.style={
        draw=gray!60, 
        rounded corners=2pt,
        fill=boxbg,
        inner sep=2pt, 
        align=left,
        drop shadow={opacity=0.1, shadow xshift=0.5pt, shadow yshift=-0.5pt}, 
        font=\sffamily\tiny 
    },
    promptstyle/.style={
        draw=gray!50,
        dashed,
        fill=yellow!5,
        rounded corners=3pt,
        font=\rmfamily\fontsize{6pt}{7pt}\selectfont, 
        align=left,
        inner sep=2pt,
        anchor=north
    },
    icon/.style={
        inner sep=0pt, 
        minimum size=0.8cm 
    },
    toplabel/.style={
        font=\rmfamily\small,
        align=center,
        anchor=south,
        text=black!80
    },
    line/.style={
        draw=linecolor,
        -Latex,
        thick
    },
    phaseheader/.style={
        draw=gray!80,
        thick,
        fill=white, 
        font=\bfseries\rmfamily\small, 
        anchor=south,
        inner sep=3pt
    }
}

\usepackage[font=bf]{caption}

\begin{document}
%

\title{\sysname: Automatically Uncovering Hidden Weaknesses in AI-Evolved Systems }



\author{
\IEEEauthorblockN{
Yajie Zhou\IEEEauthorrefmark{1},
Ao Li\IEEEauthorrefmark{2},
Ashwin Silla\IEEEauthorrefmark{2},
Zaoxing Liu\IEEEauthorrefmark{1},
Vyas Sekar\IEEEauthorrefmark{2}
}
\IEEEauthorblockA{\IEEEauthorrefmark{1}University of Maryland, College Park}
\IEEEauthorblockA{\IEEEauthorrefmark{2}Carnegie Mellon University}
}

\pagestyle{plain}

\maketitle


\begin{abstract}
The computer systems community has recently seen growing interest in AI-driven system evolution, where AI agents iteratively rewrite system programs. Frameworks such as AdaEvolve and Engram report 12--60\% score improvements over human-designed algorithms. While promising, these results raise practical concerns: AI-evolved programs may perform worse on unseen workloads or exhibit scalability regressions. Given the speed and scale of AI-generated code, we need automated mechanisms to uncover such hidden weaknesses.
We develop \sysname, which takes as input a baseline program $P$ and an AI-evolved program $P'$, and searches for valid workloads where $P'$ regresses relative to $P$ in correctness, runtime, memory usage, or output quality. To handle diverse applications and weakness types, \sysname combines deterministic workload-parameter extraction, differential oracles, and code-frequency coverage to discover distinct failures.
Across five system applications and 30 AI-evolved programs, \sysname finds 49 distinct hidden weaknesses. We also show that incorporating \sysname into the AI-driven development lifecycle can mitigate several of them. \sysname is open-sourced at \url{https://github.com/lesleychou/aichilles}.

\end{abstract}


%
\IEEEpeerreviewmaketitle

\section{Introduction}

AI is increasingly being used to optimize core system algorithms~\cite{barbarians, Barbarian-2, adaevolve, liu2026evox, adrs-db, engram, glia}. This  has generated substantial excitement in  systems research and industry.
 In essence, these frameworks follow the lead of AlphaEvolve~\cite{alphaevolve} and  cast algorithm design as an agentic code-improvement loop. For example, Google uses AlphaEvolve to improve agricultural and crop protection supply chain management~\cite{basf-alphaevolve}.
Researchers have also explored this paradigm across diverse system problems~\cite{adrs-db}, transaction scheduling~\cite{txn_scheduling}, expert parallelism load balancing~\cite{eplb}, multi-cloud job scheduling~\cite{cloudcast}, LLM prefix-cache optimization~\cite{llm_sql}, KV-cache management for model placement~\cite{prism}, and improving AI systems for industry~\cite{cocoevolve}.

Many of  these frameworks have emerged within this broader AI-driven system optimizations (\adrs) design paradigm  (e.g.,  OpenEvolve~\cite{openevolve-original}, AdaEvolve~\cite{adaevolve}, Engram~\cite{engram}). At a high level, they 
follow a similar workflow even  if they  may differ in specific design choices. To eliminate or minimize the  human effort in designing heuristics, they use an AI agent to synthesize a candidate program that is an evolved version of an original program. They include an {\em evaluator} whose function is to  score the program on a fixed set of workloads. The highest-scoring program candidates are fed back into later rounds, and this runs in an evolutionary loop.


\begin{figure}[t]
    \centering  
    \includegraphics[width=1\columnwidth]{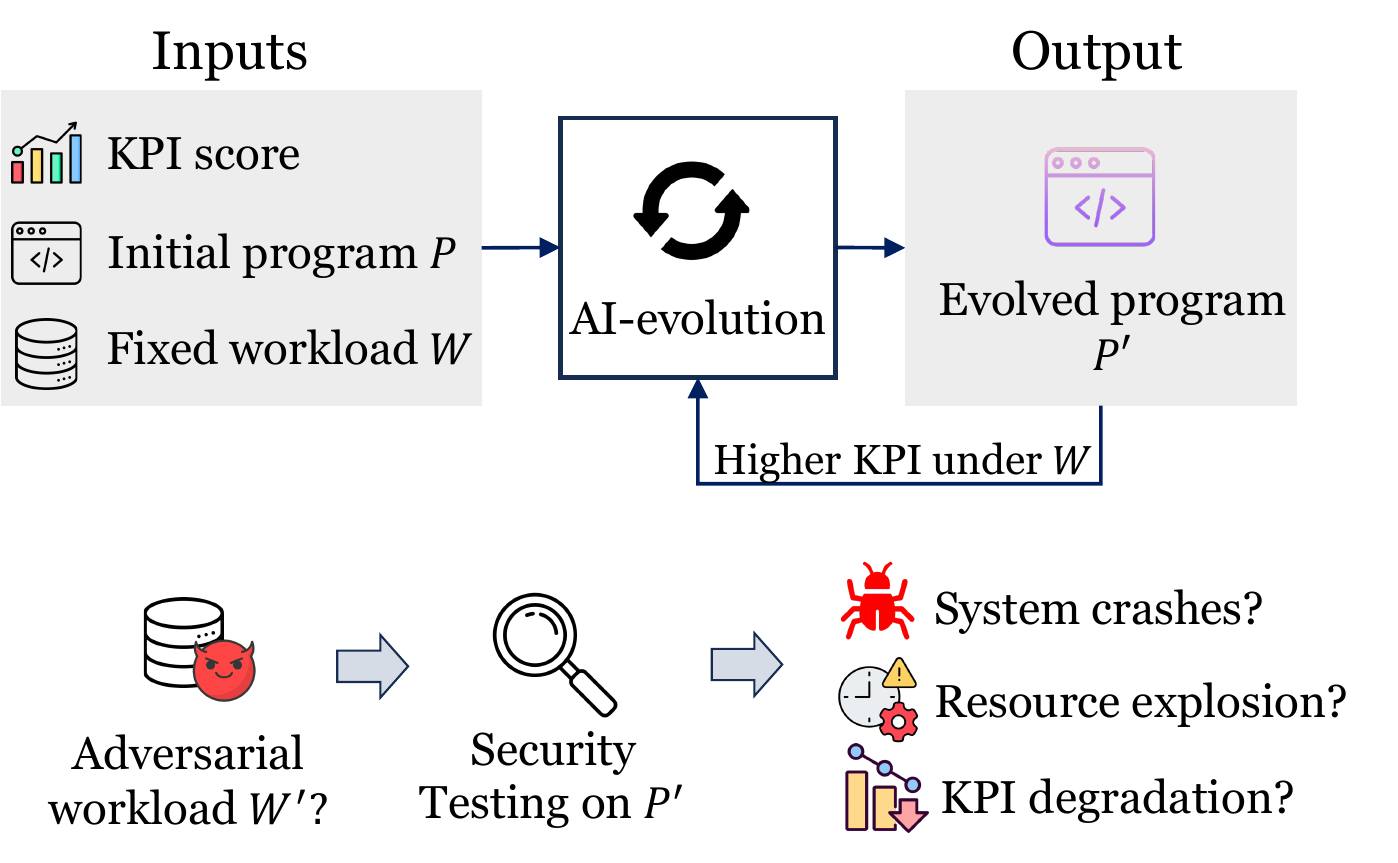} 
    \caption{AI-driven system optimization  creates an evolved program $P'$  to improve the solution quality over the initial (human-designed) program $P$. We find adversarial workloads where $P'$ instead has lower KPIs, higher resource usage, or crashes.}
    \label{fig:system_model}
    \vspace{-5mm}
\end{figure}
While this excitement to reduce human effort and improve solution optimality is understandable, we also have reason to proceed with caution when these AI-generated programs are used in critical systems applications. In particular,  AI-evolved programs can be  more complex than the original human-designed heuristics and/or overfit to the workloads used in the agentic loop. 
In operational systems, such hidden weaknesses can lead to system crashes,  hidden DoS vectors due to excessive resource consumption, and suboptimal solution quality  for future workloads. 

This concern is not merely hypothetical. We  encountered   this risk by manually inspecting the AI-evolved version of  Prism, a model-placement application for LLM serving~\cite{prism}. The original  program uses a compact greedy policy to allocate the model to the least loaded GPU. 
In contrast, we saw that the AI-evolved version replaces this single pass with a complex 
heuristic with multiple ordering strategies, multiple placement objectives, and an aggressive local search over swaps, moves, and bottleneck-GPU refinements. While this yields impressive wins  on the tested  workloads, it can  actually be  much worse on other workloads.  Furthermore, the  complexity increases compute/memory cost, which is a serious robustness concern since this program logic is on the critical path of scheduling decisions. 




Given the fast pace of AI-driven system output and code changes, we need automated ways to identify such weaknesses. However, this is challenging on four  fronts that make it difficult to  directly apply  existing techniques in program testing and fuzzing (e.g.,~\cite{fioraldi2022libafl, lemieux2018perffuzz, petsios2017slowfuzz}).
First, we need to handle a diverse set of  applications with very different input workloads and parameter spaces; e.g., one  workload may be a tuple of integer parameters such as the number of GPUs, nodes, experts, and replicas, while another may involve a sequence of read and write operations. Second, our goal is to find a {\em divergence} weakness  where  the evolved program performs worse than the reference program on the same valid input. Third,  we want to support a diverse set of  properties of interest (e.g., optimality, correctness, resource use). 
 Fourth, AI-generated code may introduce multiple points or paths of failure and we want to uncover  as many diverse root causes under a fixed time budget. 


We tackle these challenges in  designing and implementing \sysname, an agentic  system that uncovers the hidden weaknesses---the \textit{``Achilles' heel''}---of AI-evolved system programs.
\sysname treats the original human-written program $P$ as a differential oracle for the AI-evolved program $P'$. Given a valid workload, \sysname runs both programs and checks four regression types: correctness failures, execution time regressions, execution memory use  regressions, and solution optimality regressions. 

\sysname makes three design choices to make this search practical and efficient:
\begin{itemize}[leftmargin=*, topsep=2pt, itemsep=1pt, parsep=0pt, partopsep=0pt]
    \item First, we observe that the AI-evolved programs and frameworks have implicit   parameters and workload constraints hidden inside evaluators and the program itself. Strawman prompts or fuzzing approaches can miss these nuances. \sysname therefore combines deterministic parameter extraction with agent-based constraint inference. The deterministic pass extracts candidate workload parameters from the evaluator, while the agent infers valid ranges and cross-parameter constraints. 
    \item Second, given the diversity of the weakness types, we find that a monolithic approach where a single agent tries to uncover all types of weaknesses  can be  ineffective. 
    \sysname  splits the search by weakness type described above and 
    assigns  a separate subagent, which keeps the search focused and prevents it from collapsing.
    \item Third, even with the subagent approach,  we may  waste time on finding workloads that trigger the same faulty code path(s).
    To find more diverse weaknesses under a fixed time budget, \sysname uses code-frequency coverage as a behavior-diversity signal~\cite{Nguyen22-bedivfuzz}. 

\end{itemize}

Our goal is not to pinpoint flaws in a specific research prototype but to uncover systemic weakness in the AI-driven system optimizations (\adrs) design paradigm. To this end, we evaluate \sysname across
  3 diverse AI-evolution frameworks (Engram\cite{engram}, AdaEvolve\cite{adaevolve}, OpenEvolve\cite{openevolve-original}),  5 representative application  use cases (transaction scheduling~\cite{txn_scheduling}, expert-parallelism load balancing~\cite{eplb}, multi-cloud job scheduling~\cite{cloudcast}, LLM prefix-cache optimization~\cite{llm_sql}, model placement~\cite{prism}), and  2 frontier LLMs (GPT-5, Claude-Opus-4.6). 
In total,  this yields 30 AI-evolved program settings. 
 Across this spectrum, \sysname finds 49 distinct weaknesses spanning four types. The most common failures are execution-time regressions, which appear in 25 program-app instances, followed by execution-memory regressions in 11 instances, correctness weaknesses in 7 instances, and optimality regressions in 6 instances. 
These results show that hidden weaknesses are not confined to one application, model, or framework. 
No evaluated AI-evolved program family is uniformly robust under the adversarial workload search. In our experiments, Engram~\cite{engram} programs expose fewer weaknesses than AdaEvolve~\cite{adaevolve} and OpenEvolve~\cite{openevolve-original}-produced programs.
We  also compare \sysname with other testing alternatives (e.g., random fuzzing~\cite{random_fuzz}, mutational fuzzing~\cite{afl++}, property-based testing~\cite{property_fuzz}, and a baseline LLM agent). Under the same time budget, \sysname consistently finds more diverse weaknesses than these baselines.

In addition to serving as a  detection tool, we also show that \sysname can serve as an effective mitigation tool.  We find that prompt engineering alone does not prevent AI evolution from producing risky programs. However, when we add \sysname’s feedback to the AI-evolution loop and penalize candidates that expose weaknesses, the final selected programs avoid the hidden weaknesses. This robustness, however, comes at a cost: as the claimed benchmark-score improvements shrink, and in some cases the best robust program reverts closer to  the human baseline.



\para{Disclosure and Ethics.}
We disclosed the weaknesses found by \sysname and the mitigation to the authors of AdaEvolve~\cite{adaevolve} and Engram~\cite{engram}. The authors acknowledged that when an evaluator rewards only benchmark score, stronger AI evolution may exploit that objective more aggressively and reveal gaps in the evaluator. 

In some sense, our work serves as  a {\em cautionary tale} that should  temper the exuberance around \adrs. While \adrs is indeed promising, we should set realistic expectations before such solutions are ready for prime time. This paper makes the following contributions:
\begin{itemize}[leftmargin=*, topsep=2pt, itemsep=1pt, parsep=0pt, partopsep=0pt]
    \item We identify hidden regressions as security risks in AI-driven system optimization, where AI-evolved programs improve evaluator scores but may lose robustness on unseen adversarial workloads.
    \item We propose a weakness taxonomy for comparing AI-evolved programs against the original program, covering hidden weaknesses w.r.t optimality, resource use, and system crashes.
    \item We design \sysname, an agentic weakness-finding workflow that finds more weaknesses than traditional testing approaches under the same time budget.
    \item We evaluate \sysname on 5 system applications, 3 AI-evolution frameworks, and 2 LLMs, finding 49 distinct weaknesses and presenting detailed case studies across weakness types.
    \item We integrate \sysname into the AI evolution loop  and show it improves program robustness, though this reduces the original claimed improvement. 
\end{itemize}

\section{Background and Motivation}
We first introduce the background of AI-driven system optimizations (\adrs). Then we discuss  anecdotal evidence using  one  AI-optimized program  which motivates the need for   an automated tool like \sysname.

\subsection{AI-Evolution for System Algorithms}


Efforts such as AlphaEvolve~\cite{alphaevolve} and OpenEvolve~\cite{openevolve-original} show that LLM-driven evolutionary search can discover improved algorithms and optimize critical computational infrastructure. In turn, this has inspired AI-driven system optimization as a promising paradigm for automating the design of core systems algorithms. 

Systems researchers and practitioners  are increasingly exploring LLM-driven evolution  to optimize algorithms for  compute systems~\cite{barbarians, Barbarian-2, adaevolve, liu2026evox, adrs-db, engram, glia}. This has been applied across diverse tasks such as database optimization~\cite{adrs-db}, transaction scheduling~\cite{txn_scheduling}, expert-parallelism load balancing~\cite{eplb}, multi-cloud job scheduling~\cite{cloudcast}, LLM prefix-cache optimization~\cite{llm_sql}, and KV-cache-aware model placement~\cite{prism}. 
   Traditionally, these systems relied on manually designed heuristics.  The allure of   \adrs is the reduced human effort and  promise of  more optimal solutions.


At a high level, \adrs casts system algorithm design as an iterative optimization loop: an AI agent proposes candidate programs, an {\em evaluator}  scores them on given system workloads, and high-scoring candidates are retained to guide later generations.
 Many  \adrs frameworks have emerged. 
\textbf{OpenEvolve}~\cite{openevolve-original, barbarians} uses a fixed, manually-tuned strategy that controls how aggressively programs are mutated, how many candidates are retained, and which  variants to explore.
\textbf{AdaEvolve}~\cite{adaevolve} replaces this fixed strategy with dynamic control tracking   how much each  direction improves over time, and reallocates resources toward the most productive directions, and injects higher-level algorithmic guidance when progress stalls.
\textbf{Engram}~\cite{engram} targets long-horizon evolution by splitting exploration across fresh agent contexts. It preserves progress through a persistent archive and compact research digest that carry code, results, insights, and failure diagnoses across runs.
 All of these frameworks report substantial gains. For example, AdaEvolve reports best scores 12--60\% above human state-of-the-art programs on four systems tasks~\cite{adaevolve}.

\subsection{Motivating Case Study}

 The aforementioned  tasks are often on the {\em critical path of production systems} and impact  how production systems allocate resources, schedule work, route traffic, and serve models.  
 Hence our interest in  understanding potential weaknesses of such AI-evolved programs that can impact security, performance,  and robustness. 

 Driven by this  concern, we  initially  manually inspected  the effects of  \adrs applied to 
Prism~\cite{prism}, which tackles  cache-aware model placement for LLM serving systems that rely on Key-Value (KV) caches to avoid recomputing attention over previously generated tokens~\cite{pagedattention}.
 KV caches improve inference efficiency, but they also consume  GPU memory. When many models share a GPU cluster, the serving system must decide where to place each model so that request load and KV-cache pressure are balanced.

\begin{figure}[t]
\centering
\small

\begin{codepanel}{origbg}{Original $P$: one greedy pass}
for model in sort_by_pressure(models):
    # One linear scan: choose GPU with lowest current KVPR.
    g = argmin_gpu(current_kvpr(g), feasible=True)
    place model on chosen_gpu
\end{codepanel}
\begin{codepanel}{engbg}{Engram evolved $P'$: repeated heuristic search}
for rule in placement_rules:
  for order in ordering_strategies:
    # Risk 1: every candidate placement runs expensive local search.
    placement = greedy_place(models, order, rule)
    local_search(placement)  

def greedy_place(models, order, rule):
  for model in sort_by(order, models):
    # Risk 2: choose by post-placement proxy, not the original current KV pressure (KVPR).
    g = argmin_gpu(proxy_kvpr_after(model, g), feasible=True)
    place model on chosen_gpu

def local_search(placement):
  for _ in range(25):
    for g1, g2 in all_gpu_pairs():
      for m1, m2 in candidate_swaps(g1, g2):
        # Risk 3: many swap candidates repeatedly 
        # recompute max KVPR.
        if max_kvpr_after_swap(placement, m1, m2) < max_kvpr(placement):
          swap(m1, m2)
\end{codepanel}
    \vspace{-3mm}
\caption{Engram improves Prism with a more complex algorithm, but adds regression risks under new workloads.}
\label{fig:prism-code-regression}
    \vspace{-5mm}
\end{figure}

Prism defines a balancing score  called KVPR as  a proxy for avoiding the most overloaded GPU. Specifically, it  computes each GPU's KV-cache pressure as the total request pressure of its assigned models divided by its remaining KV-cache memory, then scores a placement by the inverse of the average worst-GPU pressure across test cases.
Thus, a higher score means that the placement leaves less pressure concentrated on the most constrained GPU, which should make the serving system less likely to hit a KV-cache bottleneck and miss latency targets. 
The human-designed Prism policy uses a simple greedy rule for model placement (45 lines of code).

As a case study, we consider the optimized versions output by OpenEvolve and Engram, using   Claude Opus-4.6~\cite{opus-4.6}. 
 Both use the same KVPR score defined by the PRISM paper in their {\em evaluator} function. 
OpenEvolve~\cite{barbarians} evolves a slightly larger greedy variant that ranks models by pressure weighted by model size, and places each model using a penalized post-placement KVPR score (66 lines of code). Engram~\cite{engram} discovers a much more elaborate search-based policy that achieves a higher score on Prism's original benchmark workloads (456 lines of code). 
Figure~\ref{fig:prism-code-regression} and Figure~\ref{fig:prism-code-openevolve} show the compact logic of all 3 programs.

\begin{figure}[t]
\centering
\small
\begin{codepanel}{engbg}{OpenEvolve evolved $P'$: balance-penalized future placement}

for model in sort_by((r / slo) * size):
    # Risk 1: Small but high-pressure models may be delayed because size is multiplied in.
    for gpu in all_gpus:                   
        future_kvpr = kvpr_after_placing(model, gpu)
        # Risk 2: scans all GPUs for every candidate.
        # The placement becomes O(M*G^2), not O(M*G).
        max_other = max(kvpr(g), for g in all_gpus if g != gpu)
        # Risk 3: A GPU that minimizes final max KVPR can be rejected because the penalty makes its local future_kvpr look worse.
        score = future_kvpr * (
            1.0 + 0.1 * max(0, future_kvpr - max_other)
        )
        choose gpu with smallest score
    place model on chosen_gpu
\end{codepanel}
    \vspace{-3mm}
\caption{OpenEvolve uses a shorter greedy algorithm for Prism, but it overfits to the benchmark workloads.}
\label{fig:prism-code-openevolve}
\vspace{-5mm}
\end{figure}

At first glance, both evolved programs appear  promising as they replace a compact hand-written heuristic with sophisticated    implementations. 
However, these come with hidden weaknesses as we will see next. 


\para{Runtime regression.}
The original program $P$ uses a simple greedy policy as shown. 
Thus, each model placement requires only one scan over the GPUs, giving $\mathcal{O}(MG)$ time for $M$ models and $G$ GPUs. The {\em Engram generated program} replaces this one-pass rule with repeated heuristic search. It tries many model orders and placement rules, then runs local repair after each candidate placement. The repair step  tests whether moving or swapping models between GPUs would improve the placement, and each trial recomputes GPU pressure.  
While using Engram can improve the benchmark score, its runtime grows much faster as the number of GPUs and feasible moves increases.
The {\em OpenEvolve generated program} keeps a greedy structure but makes each GPU selection more expensive. For every candidate GPU, it computes the model's pressure after placement, then scans all other GPUs to compute a balance penalty. This adds an $\mathcal{O}(G)$ inner scan inside the original $\mathcal{O}(G)$ GPU-selection loop, increasing placement time from $\mathcal{O}(MG)$ to $\mathcal{O}(MG^2)$. 

\begin{figure}[t]
    \centering
    \begin{subfigure}[t]{0.49\columnwidth}
        \centering
        \includegraphics[width=\linewidth]{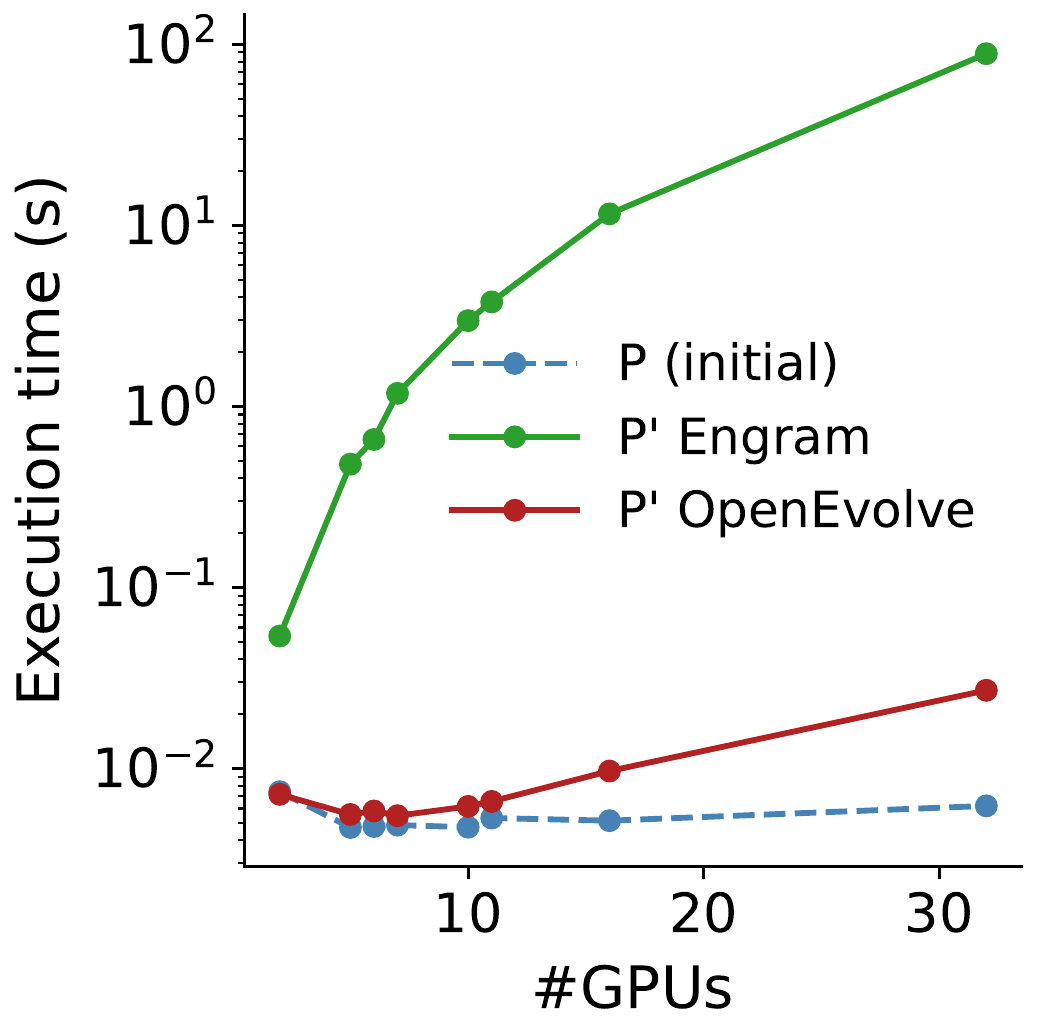}
        \caption{$P'$ shows worse and growing execution-time than $P$.}
        \label{mot:prism:time}
    \end{subfigure}
    \hfill
    \begin{subfigure}[t]{0.49\columnwidth}
        \centering
        \includegraphics[width=\linewidth]{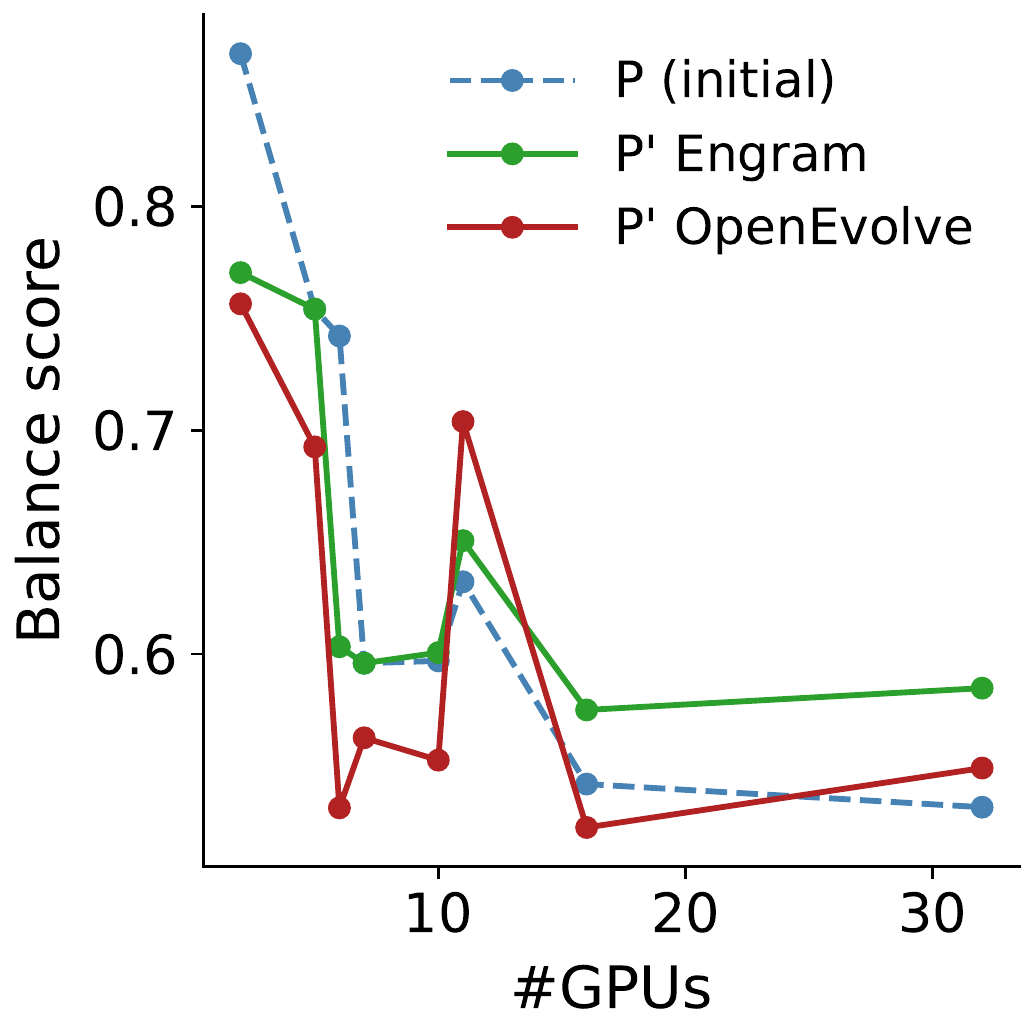}
        \caption{$P'$ shows worse performance than $P$ under some \#GPUs.}
        \label{mot:prism:perf}
    \end{subfigure}
    \vspace{-1mm}
    \caption{AI-evolved system programs expose different types of regression weaknesses when workload changes.}
    \label{fig:motivation_eplb}
    \vspace{-4mm}
\end{figure}

Such a  runtime regression can be serious when  this placement algorithm is part of an online controller.  In a shared serving cluster, placements may need to be recomputed when request rates shift, new models are added, models are removed, or GPU memory availability changes. In these settings, the scheduler must produce a new placement quickly enough for the cluster to react to the new
load. 
A policy that improves the pressure score but takes much longer to run can delay reconfiguration and leave the system operating with a stale placement while demand has already changed.
(Figure \ref{mot:prism:time} confirms the weaknesses.)


\para{Optimality regression.}
We were also concerned with 
 the risk of overfitting; i.e., the AI-evolved programs achieve higher  scores on the benchmark workloads, but can become worse  on new benchmark distributions. Figure~\ref{mot:prism:perf} confirms this regression.

 In particular, the {\em Engram generated program} expands the search space by trying many alternative placements, but each step is still driven by a local acceptance rule. A move or swap is kept only if it immediately improves Engram's internal pressure estimate. This can trap the search in a locally attractive placement: one that looks better after a single move, but blocks a better final assignment. 
 Similarly, the \textbf{\em OpenEvolve generated program} improves the benchmark score because its ``proxy'' captures patterns that are specific to the benchmark workloads, such as prioritizing large high-pressure models and avoiding GPUs whose estimated pressure would rise sharply after placement. However, these  add (possibly incorrect) assumptions that are not part of the original objective; e.g.,    model size should directly affect placement priority, and that a GPU with high projected pressure should be avoided even when using it temporarily could lead to a better overall placement. On workloads with many small high-pressure models, or workloads where a good solution requires temporarily using a pressured GPU, this proxy can misrank candidate GPUs and produce worse worst-GPU pressure than the original greedy policy.

\para{Summary.} Our manual analysis  suggests  that the AI-evolved program is not always better than the original human designed program. AI can replace a simple  algorithm with a more complex procedure that: (1) ``reward hacks'' on a small set of benchmark workloads and (b) ignores  scalability concerns. 
In this sense, the original human program may be less aggressive but is actually more robust in the face of uncertainty: it sacrifices some  optimality to maintain a stable solution.

\section{System Overview}
Our manual  analysis provides anecdotal evidence of  hidden weaknesses in \adrs. However, given the  diversity of  \adrs usage, we need new tools to  automatically find such hidden weaknesses across diverse AI-evolved system programs. We envision that such a tool becomes an indispensable part of the future CI/CD tooling for \adrs. 

\subsection{Problem Definition}
\label{sec:design}

As shown in Figure 1, the  \adrs workflow  takes a human-designed candidate program $P$, a set of workloads $W$, and an evaluator as input. It produces an AI-evolved  program $P'$ such that, for every workload $w\in \mathcal{W}$, $P'(w)$ outperforms $P$ according to the  evaluator.

Given a workload $w$, we run both programs and compare their behavior. On a workload $w \in \mathcal{W}$, an execution returns either a metric tuple or an abnormal outcome:
\[
P(w)\;\rightarrow\; \langle q,\, t,\, m \rangle \;\mid\; \bot
\]
\[
P'(w)\;\rightarrow\; \langle q',\, t',\, m' \rangle \;\mid\; \bot
\]
where $q$ is solution quality, $t$ is wall-clock execution time, $m$ is peak memory usage, and $\bot$ denotes abnormal termination.
\begin{table}[t]
  \centering
  \small
  \begin{tabular}{c p{0.58\columnwidth}}
  \toprule
  \textbf{Weakness Type} & \textbf{Weakness condition} \\
  \midrule
  \textsc{Correctness}
    & $\exists w \in \mathcal{W}: P'(w) \rightarrow \bot \wedge P(w) \not\rightarrow \bot$ \\[4pt]
  \textsc{Scalab\textsubscript{time}}
    & $\exists w \in \mathcal{W}: t' > F_t \cdot t$ \\[4pt]
  \textsc{Scalab\textsubscript{mem}}
    & $\exists w \in \mathcal{W}: m' > F_m \cdot m$ \\[4pt]
  \textsc{Optimality}
    & $\exists w \in \mathcal{W}: q' < q$ \\
  \bottomrule
  \end{tabular}
  \vspace{-1mm}
  \caption{Weakness taxonomy and conditions.}
  \label{tab:risk_type}
  \vspace{-4mm}
\end{table}

\para{Weakness definition.} We define a {\em hidden weakness}  in terms  of  a \emph{divergence violation} between these two programs. That is,  there exist valid inputs on which $P'$ behaves worse than $P$ under a security- or system-relevant metric.  We focus on four types of weaknesses in this work.  Table~\ref{tab:risk_type} summarizes the corresponding divergence checks.

\begin{itemize}[leftmargin=*, topsep=2pt, itemsep=1pt, parsep=0pt, partopsep=0pt]
    \item \textit{Correctness failure $\bot$.}
    $P'$ fails to execute successfully on a workload where $P$ succeeds, e.g., by raising an exception, or returning an invalid result.

    \item \textit{Scalability-Time $t$.}
    $P'$ has a time-scalability weakness if, on the same workload, its execution time is at least a user-defined factor $F_t$ larger than $P$.

    \item \textit{Scalability-Memory $m$.}
    $P'$ has a memory-scalability weakness if, on the same workload, its peak memory use is at least a user-defined factor $F_m$ larger than $P$.

    \item \textit{Optimality regression $q$.}
    Quality checking function returns a lower score for $P'$ than for $P$.
\end{itemize}

\para{Requirements.}
Detecting the weaknesses above imposes four  design requirements.
First, we want a \emph{general} design that can work with heterogeneous  target programs that  differ in how they
consume inputs. 
Second, it must have high  \emph{coverage across weakness types} and  expose many diverse 
weakness types in Table~\ref{tab:risk_type}, not only crashes or only
performance regressions.
Third, the discovered weaknesses must be \emph{discriminative}: an
adversarial workload should reveal a significant gap between $P$ and
$P'$, rather than being uniformly hard for both programs.
Fourth, we want \emph{diversity} to uncover 
as many distinct weaknesses as possible, instead of repeatedly
triggering the same failure mode.


\subsection{Design Choices}
\label{sec:solution_overview}

Traditional bug-finding techniques such as fuzzing~\cite{afl, afl++, fioraldi2022libafl, lemieux2018perffuzz, petsios2017slowfuzz, spider_leo} and symbolic
execution~\cite{cadar08-klee, Cadar08-exe} are effective at uncovering specific classes of weaknesses
(e.g., crashes and performance degradation), but they typically require
substantial manual effort to construct the execution environment and
testing harness for each target~\cite{green2022graphfuzz,
Domagoj19-fudge}. This per-program effort does not scale to the
diversity of AI-evolved programs we target: our evaluation alone spans
3 AI-evolving frameworks, 2 models, and 6 applications, yielding 30 distinct
programs, each with its own input format and execution setup. 
{\sysname} therefore adopts an \emph{agentic} approach, using an LLM agent to
automate the setup and search that traditional techniques leave to
human experts.

However, naively directing an LLM agent to identify weaknesses in AI-evolved algorithms fails to meet the requirements outlined above.  The workload search space is vast and often subject to complex, application-specific constraints; an unguided agent may waste its budget generating invalid workloads that violate these constraints (see \S\ref{sec:design:schema}). Moreover, a single monolithic agent tends to follow the path of least resistance: once it discovers a crash-inducing workload, it gravitates toward minor variations of the same crash, leaving scalability and optimality regressions largely unexplored (see \S\ref{sec:eval:efficiency}).

To address these issues, {\sysname} makes the following design choices:

\para{C1: Semantic-aware workload space inference.} {\sysname} adopts a two-step approach to enforce a comprehensive and semantic aware workload exploration. First, {\sysname} parses the evaluator and workload generator to extract all variables that are related to the application workload.
Second, {\sysname} explicitly prompts the agent to infer the constraints among these parameters and turns them into a workload grammar. The parser pass makes parameter discovery deterministic, while the agent supplies the semantic reasoning needed to recover application-specific validity rules. 


\para{C2: Weakness type-specific search agents.}
{\sysname} splits the search by weakness type. Each sub-agent is given one target---correctness, scalability-time, scalability-memory, or optimality---and mutates workloads to increase evidence for that target. The candidate workload is then executed on both the original program $P$ and the AI-evolved program $P'$, and the result is checked against the corresponding weakness condition. This separation keeps the search focused while keeping validation grounded in program behavior rather than the agent's own judgment.


\para{C3: Divergence-guided optimization.}
{\sysname} guides the search by the divergence between $P$ and $P'$. For each valid workload $x$, {\sysname} compares the two executions under the target weakness metric and prioritizes workloads where $P'$ behaves worse than $P$. For scalability, this means larger time or memory growth in $P'$. For optimality, it means a lower score from $P'$. For correctness, it means $P'$ fails on a workload that $P$ handles. 



\para{C4: Execution trajectory as a proxy for workload diversity.} 
{\sysname} uses execution trajectory as a proxy for behavioral diversity. For each workload, it records how often each line of $P'$ executes and represents it as a vector. A candidate is prioritized when its trajectory differs from those already explored. This steers the search toward new program behavior rather than surface-level input changes, and it also helps group repeated witnesses during root-cause analysis.

\begin{figure}[t]
    \centering  
    \includegraphics[width=1\columnwidth]{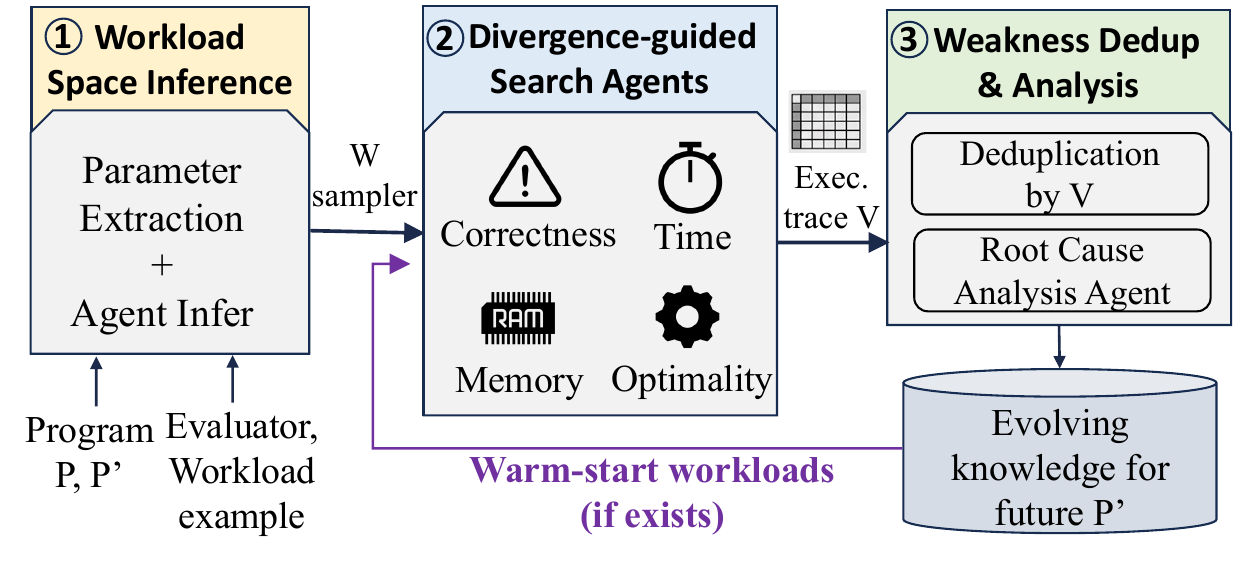} 
    \caption{\sysname Design Overview}
    \label{fig:design_overview}
    \vspace{-2mm}
\end{figure}

\section{Detailed Design}
\label{sec:design}
At the core, \sysname has three stages (Figure~\ref{fig:design_overview}):
\begin{itemize}[leftmargin=*, topsep=2pt, itemsep=1pt, parsep=0pt, partopsep=0pt]
\item 
First, \sysname infers the workload space by combining deterministic parsing with an AI agent. The parser extracts workload parameters, and the agent infers application-specific constraints to form a valid workload grammar.

\item Second, \sysname runs divergence-guided search with one agent per weakness type. Each agent searches for valid workloads where the AI-evolved program $P'$ is worse than the original program $P$ under its target metric.  \sysname uses the execution trajectory of $P'$ as a proxy for behavioral diversity and prioritizes workloads that exercise different program behavior.

\item Third, \sysname summarizes the discovered workloads into distinct weaknesses and uses an AI agent to explain their root causes. It also reuses previously found adversarial workloads as warm-start seeds when testing another evolved program for the same application.
\end{itemize}

\subsection{ Workload  Inference}
\label{sec:design:schema}
\begin{figure}[t]
    \centering  
    \includegraphics[width=0.7\columnwidth]{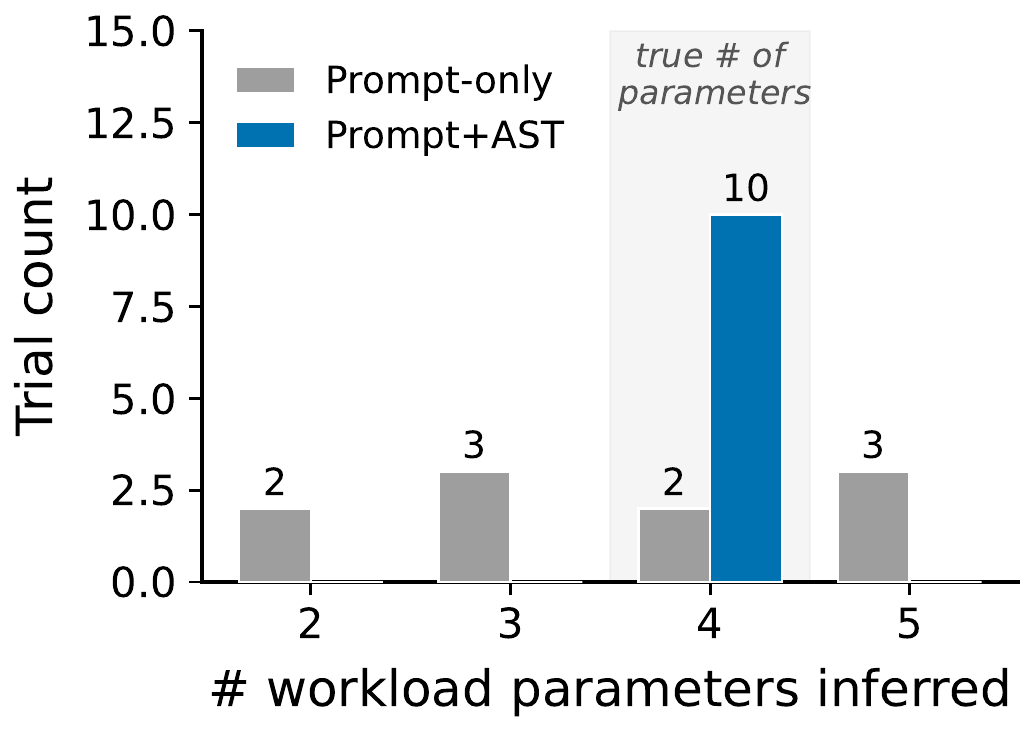} 
    \caption{Compared with \sysname's approach, Prompt-only agents produce inconsistent workload-parameter inference (example with Prism).}
    \label{fig:design:param_infer}
    \vspace{-2mm}
\end{figure}
For each AI-evolved system application, \sysname first discovers the workload space that the search is allowed to explore. In our targeted \adrs applications~\cite{skydiscover}, this space is not exposed as a clean input grammar. Workload knobs may appear in the application-specific program evaluator (that computes the program quality score), helper functions, or constants. And some validity rules are only implied by the program logic.
For example, in Prism~\cite{prism}, model placement is constrained by GPU memory capacity. For workload parameters, an individual model must fit on one GPU, so \texttt{model\_size\_max} must be below the GPU memory limit. If the parameter search ignores these constraints, it may generate workloads that crash $P'$ but are actually false positives.

A strawman approach is to directly prompt an agent to extract workload parameters and constraints for each application. We find this unreliable: across runs, the agent may identify different parameters, depending on how it interprets the application semantics. Figure~\ref{fig:design:param_infer} shows this instability in Prism, where different Opus-4.6~\cite{opus-4.6} agent trials recover very different numbers of parameters.

\sysname therefore separates deterministic parameter extraction from semantic interpretation. It first parses the evaluator using Python AST and extracts concrete workload-related values, including module-level constants, function default arguments, and literal values passed into function calls. This step does not try to understand the full application. Its purpose is to produce a stable set of candidate parameters and example values for the agent to inspect.
\sysname then prompts an agent to infer the workload grammar from these candidates and the application source files. For applications such as transaction scheduling~\cite{txn_scheduling}, the grammar can also include constraints from file metadata, such as valid column names in the workload file (see prompts in Appendix~\ref{prompt:param_infer}).

The output of this stage is a workload sampler that generates workload parameters $w$. \sysname validates the sampler by running generated workloads on the reference program $P$. If $P$ fails on $w$, \sysname uses the failure message to revise the sampler. We empirically find this validation process needs at most one revision to produce a valid workload sampler.

\subsection{Divergence-guided Search per Weakness Type}
\label{sec:design:search}
\begin{figure}[t]
    \centering  
    \includegraphics[width=1\columnwidth]{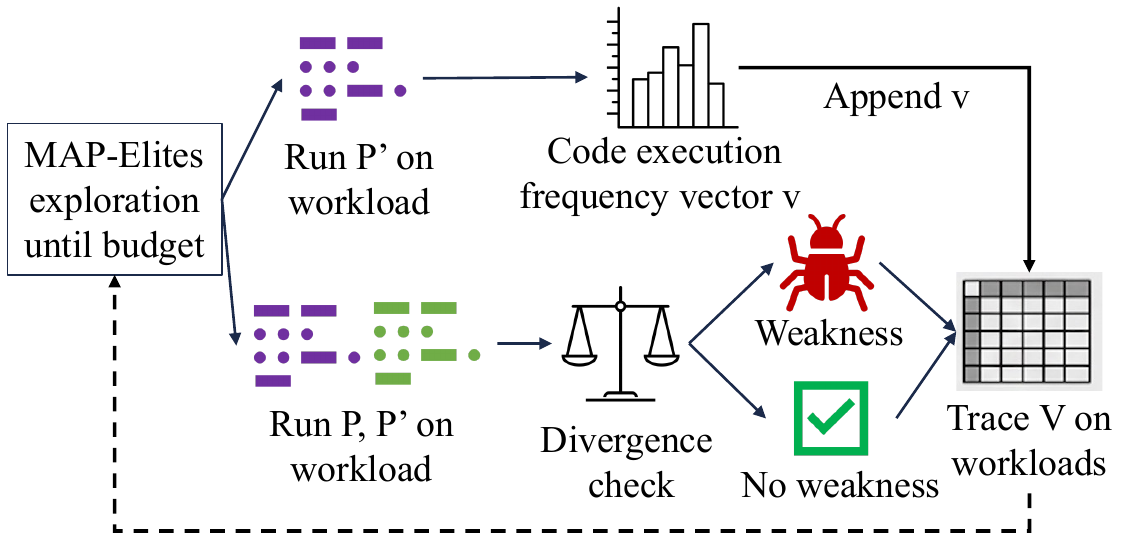} 
    \caption{Divergence-guided weakness search.}
    \label{fig:design:agent2}
    \vspace{-2mm}
\end{figure}

After obtaining a valid workload sampler, \sysname searches for workloads where $P'$ regresses relative to $P$. The first challenge is to cover different weakness types efficiently.
A strawman approach is to use one monolithic agent to search for all possible weaknesses. However, we find this unreliable. For instance,  once an  agent finds a crash-inducing workload, it  tends to  propose nearby crash-inducing  cases and misses scalability or optimality weaknesses. This behavior is consistent with prior evidence that LLMs can struggle to use long prompts effectively~\cite{lost-in-the-middle}. 

\begin{figure}[t]
    \centering
    \begin{subfigure}[t]{0.48\columnwidth}
        \centering
        \includegraphics[width=\linewidth]{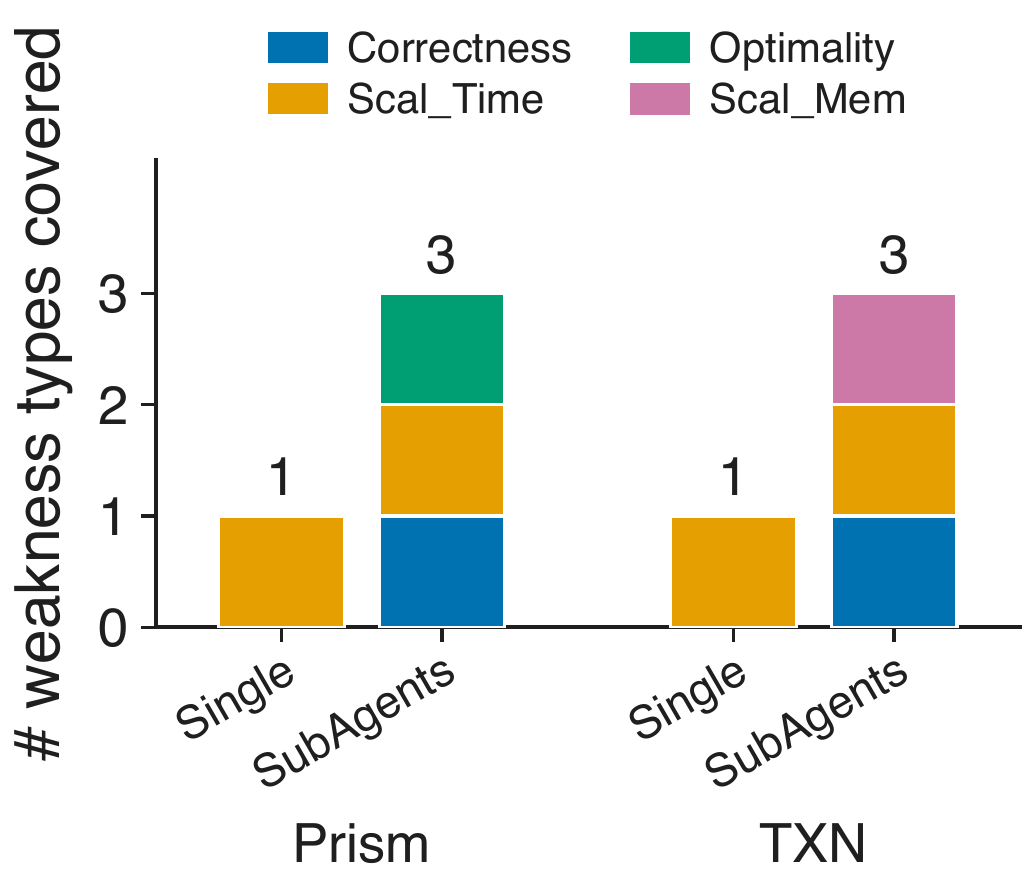}
        \caption{\sysname's sub agents find more diverse weakness types than a single agent.}
        \label{fig:design:sub_agent}
    \end{subfigure}
    \hspace{0.02\columnwidth}
    \begin{subfigure}[t]{0.47\columnwidth}
        \centering
        \includegraphics[width=\linewidth]{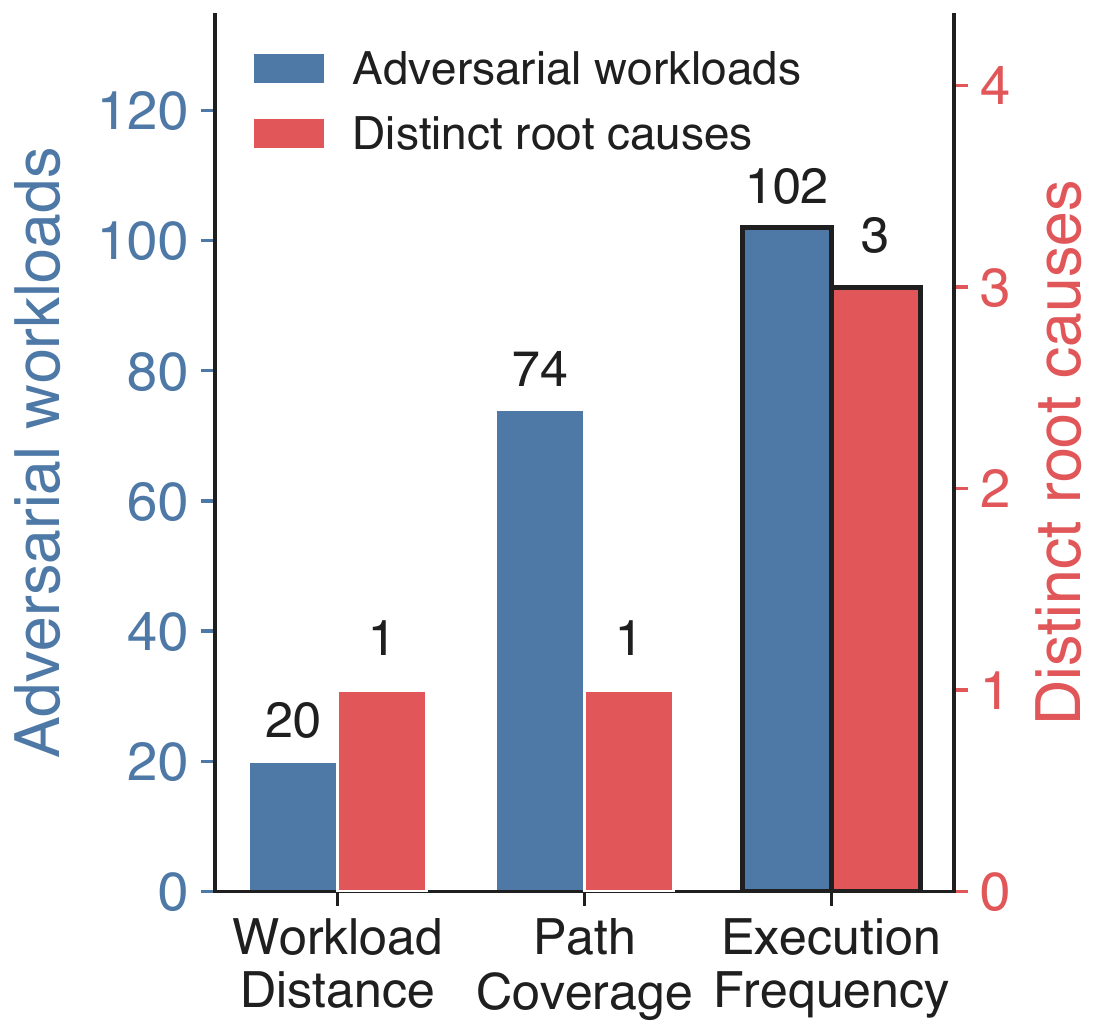}
        \caption{Code execution frequency is a better diversity proxy than alternatives.}
        \label{fig:design:proxy}
    \end{subfigure}
        \vspace{-2mm}
    \caption{Design choices for workload search.}
    \label{fig:design:search}
    \vspace{-2mm}
\end{figure}

\sysname therefore uses one sub-agent per weakness type. Each sub-agent is prompted with the corresponding divergence definition in Table~\ref{tab:risk_type}, so it focuses on workloads likely to expose that specific regression. This design keeps the search targeted: correctness agents look for failures of $P'$ that do not occur in $P$, scalability agents look for time or memory gaps, and optimality agents look for quality score regressions. Figure~\ref{fig:design:sub_agent} shows that a single agent only finds one weakness type, while \sysname's type-specific agents cover three weakness types on both Prism and TXN.

The second challenge is finding diverse instances within each weakness type.   
 A seemingly natural but incorrect  choice to measure diversity is to consider some  {\em distance metric over the  input workload}; e.g.,   Euclidean distance between workload-parameter vectors. However,  this is misleading in our setting.  Two workloads can be far apart in parameter space but still execute the same code path and trigger the same root cause. Conversely, two nearby workloads may reach different branches and expose different failures. This wastes the testing budget and can overstate the number of distinct weaknesses.  
 

We  need a diversity metric that reflects program behavior, not just input variation. To this end, \sysname  uses the execution trajectory of $P'$ as a proxy for workload diversity. For each workload, \sysname records which parts of $P'$ execute and tracks their execution frequencies using a fixed-size counter map. This captures not only whether a workload reaches a code region, but also how heavily it exercises loops, branches, and helper functions.
Other fuzzing approaches also use path coverage as a diversity signal~\cite{wang2020not}. 
Figure~\ref{fig:design:proxy} compares three choices: workload distance, path coverage, and execution frequency. We find that execution frequency exposes the largest number of distinct weaknesses, because many AI-evolved system program regressions are not caused by reaching a new branch alone. They often come from repeatedly exercising the same logic under different load conditions, such as deeper search, larger intermediate structures, or repeated placement updates. The execution frequency therefore provides a finer signal than path coverage while staying closer to root-cause behavior than raw workload distance.

Combining these two key ideas, Algorithm~\ref{alg:diverse_workload_search} shows how \sysname searches for adversarial workloads. For each weakness type $\tau \in \mathcal{T}$ and given application, \sysname starts from workloads sampled from the grammar and previously identified workloads. A type-specific agent mutates a seed workload into new candidates. \sysname runs each candidate on $P'$ to record its behavior and execution trajectory, skips candidates that repeat explored behavior, and then runs the remaining candidates on $P$. If $P'$ is worse than $P$ under the target weakness condition, the workload is added to $\mathcal{W}_{\tau}$.

\sysname uses MAP-Elites~\cite{map-elites} to keep the search robust  and diverse. MAP-Elites is a standard Quality-Diversity optimization algorithm: it divides explored behavior into coarse cells, and each cell keeps the best candidate found for that behavior. 
In \sysname, the cell is computed from the execution trajectory of $P'$. Within each cell, \sysname keeps the workload with the largest divergence score $d$. The per-type archive $\mathcal{A}_{\tau}$ stores useful seeds for the current weakness type, while the global archive $\mathcal{A}$ tracks behavior explored across all weakness types (Algorithm~\ref{alg:map_elites_update}). This prevents the search from only chasing one repeated failure and helps it keep workloads that expose different program behaviors.

\begin{algorithm}[t]
\caption{Divergence-guided Workload Search}
\label{alg:diverse_workload_search}
\begin{algorithmic}[1]
\Require Initial program $P$, evolved program $P'$, workload grammar $G$, weakness types $\mathcal{T}=\{c,t,m,q\}$, search budget $B$
\Ensure Adversarial workload sets $\{\mathcal{W}_{\tau}\}_{\tau \in \mathcal{T}}$

\State Initialize global archive $\mathcal{A}$, abnormal-workload set $\mathcal{C}$, and adversarial workload sets $\{\mathcal{W}_{\tau}\}$
\State Set per-type budget $B_{\tau} \gets \lfloor B/|\mathcal{T}| \rfloor$

\ForAll{$\tau \in \mathcal{T}$}
    \State $\mathcal{A}_{\tau} \gets \textsc{WarmStart}(G,\tau)$
    \While{$B_{\tau} > 0$}
        \State $w \gets \textsc{SampleSeed}(\mathcal{A}_{\tau})$
        \State $S \gets \textsc{AgentMutate}(w,G,\tau,\mathcal{C})$

        \ForAll{$\hat{w} \in S$}
            \State $(q',t',m',v') \gets \textsc{RunWithTrajectory}(P',\hat{w})$; \quad $B_{\tau} \gets B_{\tau}-1$
            \If{$\textsc{Skip}(\hat{w},v',\mathcal{A},\mathcal{C},\tau)$}
                \State \textbf{continue}
            \EndIf

            \State $(q,t,m) \gets \textsc{Run}(P,\hat{w})$
            \State $(\mathcal{W},\Delta) \gets \textsc{CheckWeaknesses}(q,t,m,q',t',m',\hat{w},v')$
            \State $\textsc{MapElitesUpdate}(\mathcal{A}_{\tau},\hat{w},v',\Delta[\tau])$
            \State $\textsc{MapElitesUpdate}(\mathcal{A},\hat{w},v',\max_{\eta\in\mathcal{T}}\Delta[\eta])$
        \EndFor
    \EndWhile
\EndFor

\State \Return $\{\mathcal{W}_{\tau}\}_{\tau \in \mathcal{T}}$
\end{algorithmic}
\end{algorithm}
\begin{algorithm}[t]
\caption{MAP-Elites Updating Search}
\label{alg:map_elites_update}
\begin{algorithmic}[1]
\Require Archive $\mathcal{A}$, workload $w$, execution trajectory $v$, divergence score $d$
\Ensure Updated archive $\mathcal{A}$

\State $c \gets \textsc{Cell}(v)$ \Comment{Map execution behavior to an archive}

\If{$c \notin \mathcal{A}$}
    \State $\mathcal{A}[c] \gets (w, v, d, 0)$
    \Comment{Store first workload}
\ElsIf{$d > \mathcal{A}[c].d$}
    \State $\mathcal{A}[c] \gets (w, v, d, \mathcal{A}[c].visits)$
    \Comment{Keep the strongest regression in this cell}
\EndIf

\State \Return $\mathcal{A}$
\end{algorithmic}
\end{algorithm}

\subsection{Weakness Summarization}
\label{sec:design:report}
\begin{figure}[t]
    \centering  
    \includegraphics[width=0.8\columnwidth]{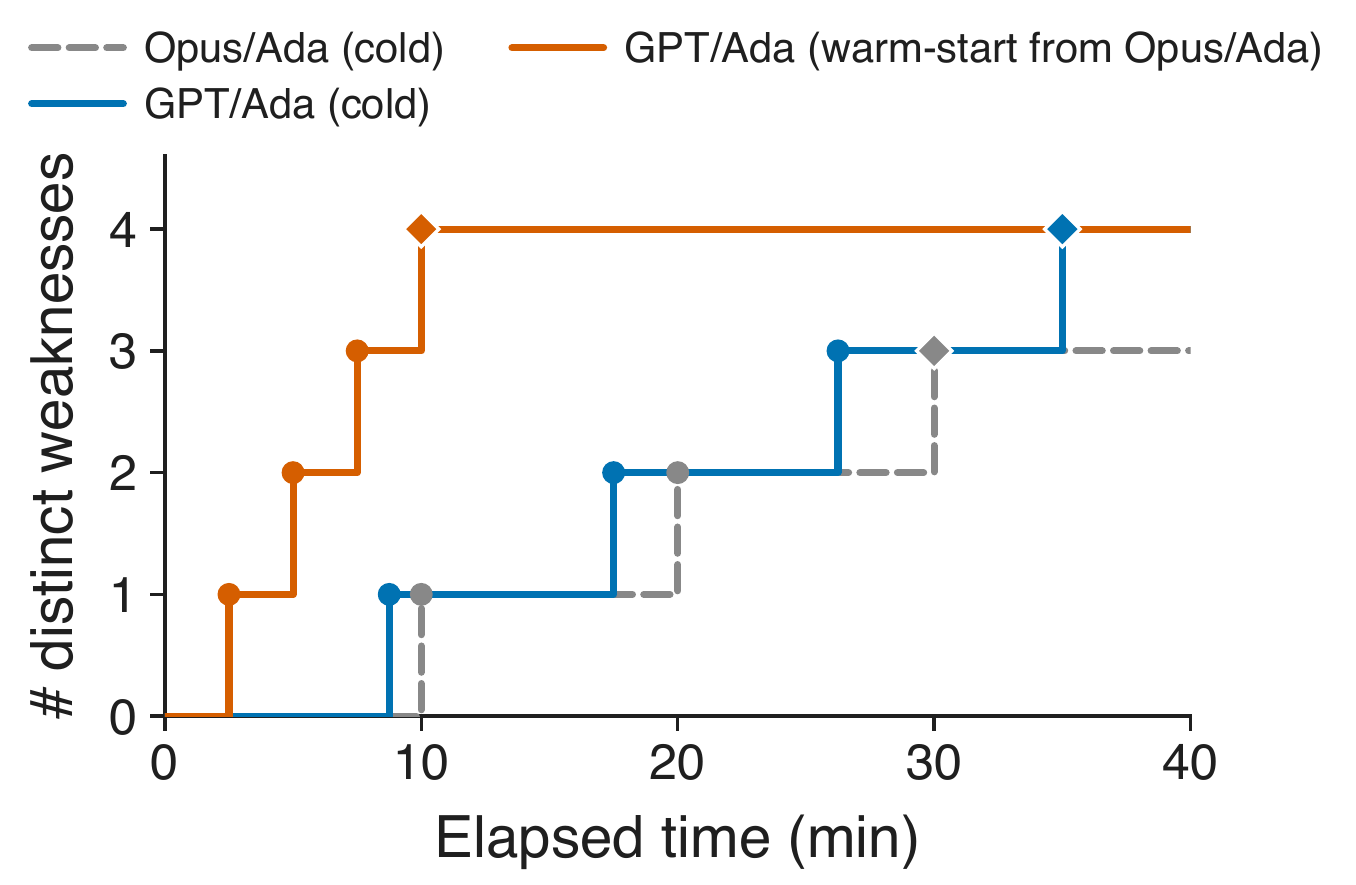} 
     \vspace{-2mm}
    \caption{With warm-start, \sysname finds weaknesses in GPT/AdaEvolve-generated Prism $P'$ more efficiently.}
    \label{fig:design:warm_start}
    \vspace{-4mm}
\end{figure}
After \sysname finds adversarial workloads,  human (or agentic) reviewers still need a concise report of distinct root causes. Execution-trajectory diversity reduces repetition during search, but it does not ensure  unique  root causes; e.g.,   two workloads may execute different line frequencies yet fail because of the same flawed function.

\sysname therefore deduplicates adversarial workloads by the execution behavior of $P'$. For each workload, it uses the collected code-frequency vector to find the lines that dominate the run, then maps those lines back to functions using AST source ranges. If a line belongs to nested functions, \sysname assigns it to the innermost function, which best captures the local code responsible for the behavior. The function with the largest accumulated weight becomes the workload's \emph{trigger function}.
\sysname groups adversarial workloads that share the same trigger function and keeps the workload with the largest divergence score as the representative example. This gives each group a concrete workload and measured regression, while the group size shows how broadly the same root cause appears.

Finally, \sysname asks an agent to write a short root-cause explanation for each group. The prompt includes the representative workload, weakness type, divergence score, trigger function, relevant source lines in $P'$, and the corresponding behavior of $P$. The agent does not decide whether a weakness exists; it only explains why that region of $P'$ causes the observed regression compared with $P$ (see prompts in Appendix~\ref{prompt:root_cause}).

\para{Evolving knowledge across $P'$.}
After testing one evolved program $P'$, \sysname stores useful search results for later evolved programs from the same application. The knowledge base contains three types of workloads: adversarial workloads that satisfy a weakness condition, high-divergence workloads that do not cross the weakness threshold, and summary statistics about which weakness types were observed. The first group captures regressions that may recur in another $P'$. The second group captures near misses: workloads that stress the current $P'$ and may become adversarial for a different evolved implementation.

When \sysname tests a new $P'$ for the same application, it replays these stored workloads before starting the mutation loop. Each workload is re-evaluated against the new $P'$, so \sysname does not assume that a previous weakness still exists. Even when a stored workload no longer satisfies a weakness condition, its execution trajectory can still seed the MAP-Elites archive with useful program behavior. Figure~\ref{fig:design:warm_start} shows that this warm start helps \sysname discover distinct weaknesses faster and begin the search from a more diverse set of behaviors.

\section{Evaluation}
\label{sec:eval}
We study three questions to evaluate \sysname. 

\begin{itemize}[leftmargin=*, topsep=2pt, itemsep=1pt, parsep=0pt, partopsep=0pt]
    \item \textbf{RQ1: Can \sysname find distinct weaknesses efficiently?}
    Under the same budget, we compare \sysname with existing program testing approaches and measure if it finds more distinct weaknesses. (\S\ref{sec:eval:efficiency})
    
    \item \textbf{RQ2: What hidden weaknesses does \sysname find in AI-evolved programs?}
    We measure how many weaknesses \sysname finds, their root causes, and what risks they imply for applying AI-evolved programs in practical systems with case studies. (\S\ref{sec:eval:bug_summary}, \S\ref{sec:eval:case_study})

    \item \textbf{RQ3: If we include \sysname in the evaluation loop, can it mitigate the weaknesses?}
    We test whether prompt engineering and evaluator patching, when added to the AI-evolution loop, prevent the evolved program from introducing hidden regressions. (\S\ref{sec:eval:mitigation})

\end{itemize}

\para{Setup.}
We evaluate \sysname on 3 AI-evolving frameworks released by prior work: OpenEvolve~\cite{skydiscover}(version \texttt{de8086f}), AdaEvolve~\cite{skydiscover}(version \texttt{de8086f}), and Engram~\cite{engram}(version \texttt{5295858}). We test each framework with two frontier LLMs: GPT-5~\cite{gpt-5} and Claude Opus-4.6~\cite{opus-4.6}.
We then select the \emph{highest-scoring program} under the original evaluator -- the same artifact that prior work cites as evidence of human-competitive performance.

\begin{figure*}[t]
    \centering  
    \includegraphics[width=\linewidth]{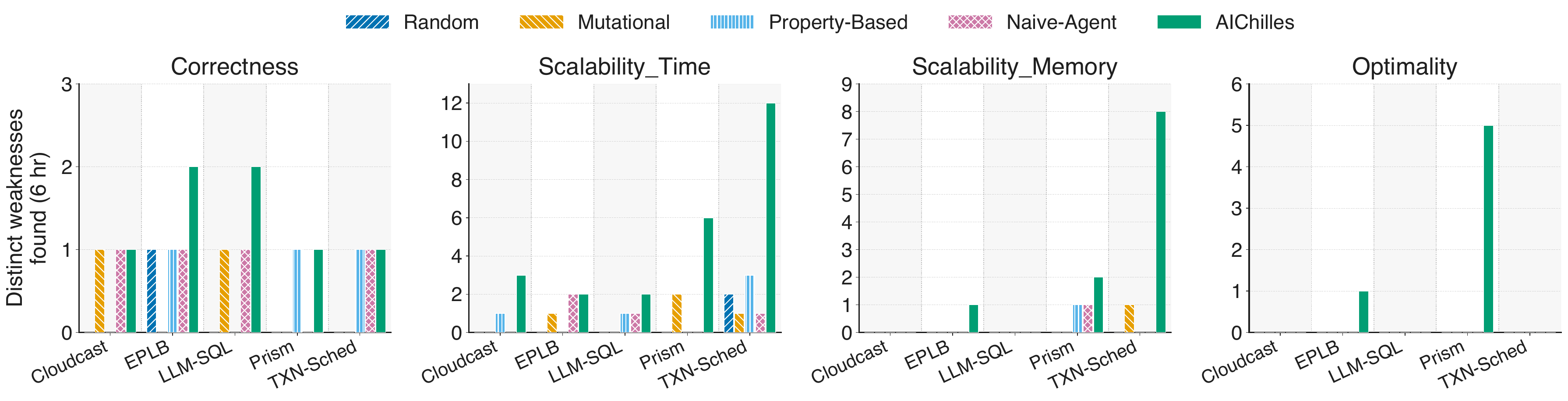} 
        \vspace{-6mm}
    \caption{Compared with baselines, \sysname finds more distinct weaknesses within a 6-hour budget.}
    \label{fig:eval:pass_at_t}
    \vspace{-5mm}
\end{figure*}
\para{Applications.}
We evaluate 5 system applications collected from prior work~\cite{adaevolve, skydiscover}.

\begin{itemize}[leftmargin=*, topsep=2pt, itemsep=1pt, parsep=0pt, partopsep=0pt]
    \item \textit{Transaction scheduling (TXN-Sched)~\cite{txn_scheduling}} optimizes the execution order of database transactions that read and write shared keys.
    It reduces lock conflicts and minimizes total makespan on transactional workloads.
    The evaluator checks whether an evolved scheduler can preserve conflict constraints while compressing completion time.

    \item \textit{Expert parallelism load balancing (EPLB)~\cite{eplb}} targets load imbalance in Mixture-of-Experts inference, where popular experts can overload a subset of GPUs.
    The algorithm must decide how many replicas each expert should have and how to place them across GPUs.
    The evaluator measures load-balance quality and the runtime cost of rebalancing.

    \item \textit{Multi-cloud job scheduling (Cloudcast)~\cite{cloudcast}} studies cost-aware data transfer across multi-region and cloud.
    The evaluator validates complete, valid delivery paths for all destinations and partitions across cloud configurations, then scores the algorithm by inverse total transfer cost.

    \item \textit{LLM prefix-cache optimization (LLM-SQL)~\cite{llm_sql}} optimizes relational analytics workloads where each table row is serialized into an LLM prompt.
    The algorithm reorders rows and fields to increase shared prefixes between consecutive prompts.
    The evaluator rewards high prefix-cache hit rate while keeping the reordering algorithm itself efficient.

    \item \textit{Model placement (Prism)~\cite{prism}} places multiple LLMs onto a fixed GPU cluster under heterogeneous and bursty serving demand.
    The evaluator checks the model placement under GPU memory constraints, then scores placements by low average KV-cache pressure and success rate.
\end{itemize}

\para{Baselines.} 
We compare \sysname with four practical alternatives for testing Python programs.

\begin{itemize}[leftmargin=*, topsep=2pt, itemsep=1pt, parsep=0pt, partopsep=0pt]
      \item \textit{Random Fuzzing~\cite{random_fuzz}} uniformly samples independent workload parameters from their ranges at each
  trial, with no memory of prior results.

    \item \textit{Mutational Fuzzing~\cite{afl++}} starts from uniformly sampled workloads and keeps a corpus of promising cases. At each step, it mutates the workload with the largest performance gap: numeric parameters are changed by up to 20\% within their valid ranges, and categorical parameters are resampled with 30\% probability. The resulting workloads are ranked by performance gap for future mutation.

    \item \textit{Property-Based Testing~\cite{property_fuzz}} uses the Hypothesis library~\cite{hypothesis} to automatically generate and shrink workloads. For each workload parameter, we specify the valid values that Hypothesis may sample. The fuzzer then searches for workloads that violate the property \emph{``the AI-evolved program is never outperformed by the initial program.''}

   \item \textit{Naive-agent} puts everything in the prompt as a single weakness finding agent (see prompts in Appendix~\ref{prompt:naive_agent}). It uses the same condition to confirm weaknesses but does not include \sysname's specific design choices.
  \end{itemize}


\para{Metric.}
For each generated workload $w$, we run it on both the initial program $P$ and the AI-evolved program $P'$.
A workload is \emph{adversarial} if it exposes one or more of the target conditions: \emph{correctness}, \emph{scalability\_time}, \emph{scalability\_memory}, or \emph{optimality}.
We measure how many \emph{distinct weaknesses} a method finds under a fixed time budget $T$.
A \emph{distinct weakness} is defined as a group of adversarial workloads that satisfy a weakness condition and share the same trigger function in $P'$, which serves as a proxy for the same root cause.

\subsection{Effectiveness and Diversity}
\label{sec:eval:efficiency}
To compare weakness-finding efficiency, we measure the number of distinct weaknesses found within a six-hour budget. Each method is run five times with different random seeds, and we report the average. For fairness, all baselines use the same workload parameters recovered by \sysname's workload-space inference stage; the comparison therefore focuses on search effectiveness rather than differences in parameter discovery.

We aggregate results of 6 AI-evolved programs on each application. Figure \ref{fig:eval:pass_at_t} shows that \sysname finds more distinct weaknesses than all baselines across all four weakness types. The gap is smallest for correctness weaknesses, where simple mutations and naive LLM agents can sometimes find crashes. However, after finding one crash pattern, these methods often keep generating similar crash workloads  and miss other weakness types.

The gap becomes much larger for regressions that require comparing $P'$ against $P$. For scalability-time, \sysname finds up to 12 distinct weaknesses in TXN, while no baseline finds more than 3 across all applications. For optimality, none of the baselines finds a distinct weakness. The optimality cases are harder because the failure often depends on a specific combination of workload parameters. Changing one parameter alone may not make the program quality regression worse, and combining two promising changes may even hide the regression. This makes simple random mutation or naive-agent search less effective.

We also compare the cost of each weakness-finding method in Table~\ref{tab:search-cost}. Using Prism as an example, traditional baselines consume nearly the full six-hour budget on one CPU core, with roughly 95--100\% CPU utilization. Agent-based methods use fewer local CPU-hours but add LLM token cost.
\sysname uses more local CPU than the naive-agent baseline because it runs divergence checks and tracks execution trajectories during search. However, it uses fewer tokens since each sub-agent focuses on one weakness type, and the execution-frequency signal guides workload selection, instead of relying on the agent to reason about all four weakness types at every iteration.
Although all methods are given a six-hour budget, we find that \sysname often reaches its final set of distinct weaknesses earlier. On Prism, EPLB, and Cloudcast, \sysname converges to the reported distinct weaknesses within 20--30 minutes.

\begin{table}[t]
\centering
\small

\begin{tabular}{lcc}
\toprule
\textbf{Method} & \textbf{CPU (hours)} & \textbf{Token (\$)} \\
\midrule
Random       & 5.8 & - \\
Mutational   & 6.0 & - \\
Property     & 6.7 & - \\
Naive-agent  & 1.2 & 10.24 \\
\sysname     & 2.4 & 8.95 \\
\bottomrule
\end{tabular}
\caption{Search cost on Prism, with token cost computed using Opus-4.6 pricing~\cite{opus-4.6}.}
\label{tab:search-cost}
    \vspace{-6mm}
\end{table}

\begin{tcolorbox}[colback=green!6!white,colframe=green!50!black,sharp corners=south,boxrule=0.8pt,enhanced,width=\linewidth,arc=3pt,left=4pt,right=4pt,top=4pt,bottom=4pt]
\textbf{\textit{Takeaway 1:}} 
\textit{
\sysname finds 49 distinct weaknesses spanning all four types. 
The bottleneck of finding hidden weaknesses is not generating more workloads, but generating workloads that expose how $P'$ behaves differently from $P$. Baselines can find program crashes, but they struggle with scalability and optimality regressions because these failures depend on specific workload interactions. \sysname acts more like a human auditor, it searches for divergence directly and prioritizes workloads that exercise new behavior in $P'$.
}
\end{tcolorbox}

\subsection{Weakness Pattern Analysis}
\label{sec:eval:bug_summary}
\begin{figure}[t]
    \centering  
    \includegraphics[width=1\columnwidth]{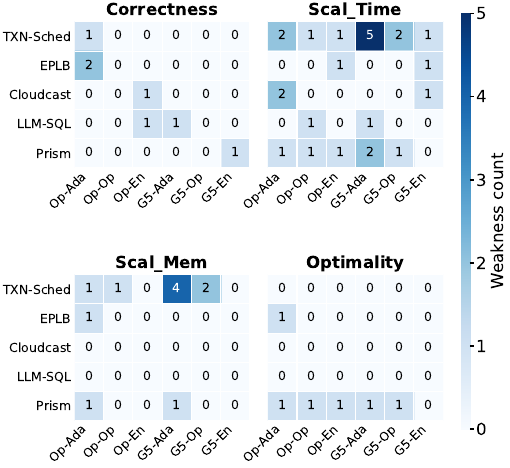}
    \vspace{-6mm}
    \caption{Weaknesses found across 5 applications, 3 AI frameworks, and 2 LLMs.}
    \label{fig:eval:heatmap}
    \vspace{-4mm}
\end{figure}

  Next we dive deeper to analyze patterns in the uncovered weaknesses.  
Figure~\ref{fig:eval:heatmap} shows details of distinct weaknesses found from \sysname on each application and each AI-evolved program.

\para{Application-level patterns.}
We see that different target  applications expose different kinds of weaknesses. \textsc{TXN-Sched} is the clearest scalability case. In this application, GPT/AdaEvolve alone produces five scalability-time weaknesses and four scalability-memory weaknesses. 
In \textsc{Prism}, all six evolved programs contain at least one weakness. It is also where optimality regressions are most concentrated. This is interesting because Prism programs can remain executable while making worse placement decisions. A bad model-placement strategy may not crash or immediately exhaust memory, but it can still produce a lower placement score by increasing KV-cache pressure. Thus, Prism exposes weaknesses that are silent quality regressions rather than obvious failures. Among all applications, \textsc{Cloudcast} and \textsc{EPLB} have fewer distinct weaknesses, and their failures are concentrated in fewer framework/model combinations.

\para{AI framework-level patterns.}
The AI-evolving framework affects not only how many weaknesses appear, but also what kind of weakness appears. Engram-generated programs have a narrower profile in our results: Claude/Engram and GPT/Engram both expose correctness and scalability-time weaknesses, but neither exposes scalability-memory or optimality weaknesses. AdaEvolve produces all four types of weaknesses.
We find that in the \textsc{TXN-Sched} cases, AdaEvolve introduces heavier optimization procedures that rewrite the whole program logic, which creates more opportunities for algorithmic blowups. Engram generated program stays closer to the original human-designed program implementation, with less aggressive rewrites. (Table \ref{tab:txn_code_compare})

\para{Model-level patterns.}
The LLM model also matters, but its effect depends more on the AI-evolving framework. Under Engram, Claude and GPT have similar weakness profiles. Under AdaEvolve, the model difference is larger. For example, Opus/AdaEvolve on \textsc{EPLB} exposes correctness, memory, and optimality weaknesses in the same application, while GPT/AdaEvolve exposes none there. On \textsc{TXN-Sched}, GPT/AdaEvolve is the worst case, with both the largest time and the largest memory weakness count.
This suggests that model choice is not an independent factor. 
The AI-evolving framework controls how much room the model can rewrite the program structure, while the model affects which concrete strategy is produced inside that program.

\begin{tcolorbox}[colback=green!6!white,colframe=green!50!black,sharp corners=south,boxrule=0.8pt,enhanced,width=\linewidth,arc=3pt,left=4pt,right=4pt,top=4pt,bottom=4pt]
\textbf{\textit{Takeaway 2:}} 
\textit{Weaknesses in AI-evolved programs depend on both the application and the evolution framework. For example, Prism exposes silent quality regressions, while TXN-Sched exposes resource blowups. Different frameworks, such as Engram and AdaEvolve, also produce different weakness profiles. Auditing should therefore account for the kinds of transformations each AI-evolving framework tends to encourage.
}
\end{tcolorbox}

\subsection{Case Studies}
\label{sec:eval:case_study}
 To help shed light on the structure of the weaknesses, next we present more in-depth case studies.

\para{Correctness weaknesses in EPLB.}
EPLB balances mixture-of-experts serving by deciding how many physical replicas each logical expert should receive. Intuitively, a heavily loaded expert should be copied more times, so that its traffic can be split across more physical experts. 
Figure~\ref{fig:eplb-code-correctness} shows  the initial program using a one-step greedy loop: before assigning each new replica, it recomputes the current load per replica, chooses the most overloaded logical expert, assigns one physical replica, and updates that expert's replica count. The AdaEvolve  program keeps the same goal but batches the assignments. In each batch, it computes normalized load once, selects a fixed \texttt{topk} list of high-load experts, and then fills several replica slots from that list.
The  crash happens because the  evolved program implicitly assumes  the batch size is not larger than the number of experts returned by \texttt{topk}. This fails when the workload has few logical experts but many physical replicas. For example, with 8 logical experts and 288 physical replicas, the final phase uses batch size 20, but \texttt{topk(min(batch\_len, num\_log))} returns only 8 experts. The  loop still iterates over 20 slots, indexes past the candidate tensor and crashes. Note that  this is a correctness failure on a valid EPLB workload. The original human-designed program avoids it because it assigns one replica at a time.


\begin{figure}[t]
\centering
\small

\begin{codepanel}{origbg}{Original: safe one-replica-at-a-time update}
for i in range(num_log, num_phy):
    redundant_indices = (weight / logcnt).max(dim=-1).indices
    phy2log[:, i] = redundant_indices
    rank[:, i] = logcnt[arangen, redundant_indices]
    logcnt[arangen, redundant_indices] += 1
\end{codepanel}
\begin{codepanel}{engbg}{Opus/AdaEvolve evolved: batched top-$k$ update}
batch_size = 20
for batch_start in range(phase2_end, num_phy, batch_size):
    batch_end = min(batch_start + batch_size, num_phy)
    batch_len = batch_end - batch_start
    normalized_load = weight / logcnt.float()
    if batch_len == 1:
        redundant_indices = normalized_load.max(dim=-1).indices.unsqueeze(-1)
    else:
        #Risk: if batch_len > num_log, this returns only num_log columns.
        _, redundant_indices = normalized_load.topk(min(batch_len, num_log), dim=-1)
        redundant_indices = redundant_indices[:, :batch_len]

    for j in range(batch_len):
        #Risk: j may exceed redundant_indices.shape
        expert_idx = redundant_indices[:, j]
        phy2log[:, batch_start + j] = expert_idx
        rank[:, batch_start + j] = logcnt[arangen, expert_idx]
        logcnt[arangen, expert_idx] += 1
\end{codepanel}
    \vspace{-2mm}
\caption{\textbf{EPLB correctness weakness}}
\label{fig:eplb-code-correctness}
    \vspace{-4mm}
\end{figure}

\begin{figure}[t]
    \centering
    \begin{subfigure}[t]{0.49\columnwidth}
        \centering
        \includegraphics[width=\linewidth]{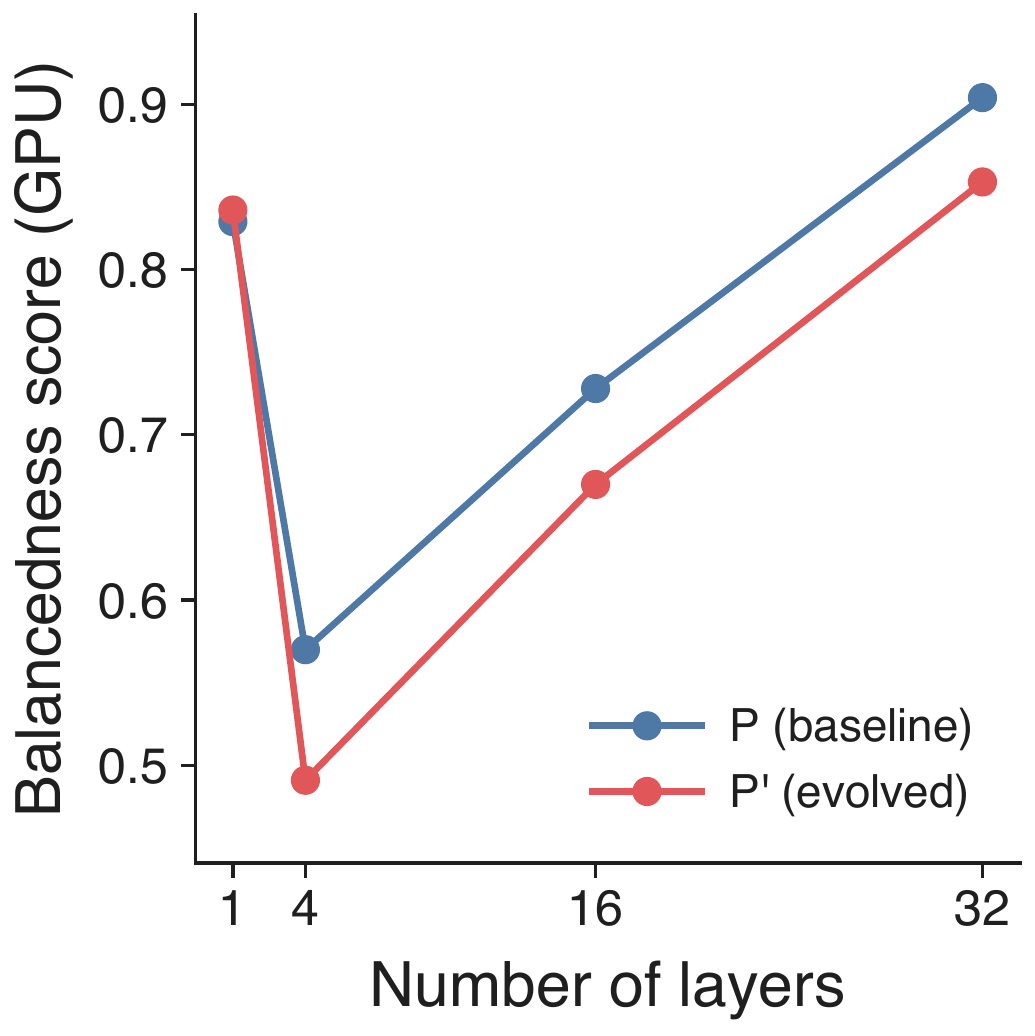}
        \caption{Score w. increasing layers}
        \label{fig:eval:eplb_layer}
    \end{subfigure}
    \hfill
    \begin{subfigure}[t]{0.49\columnwidth}
        \centering
        \includegraphics[width=\linewidth]{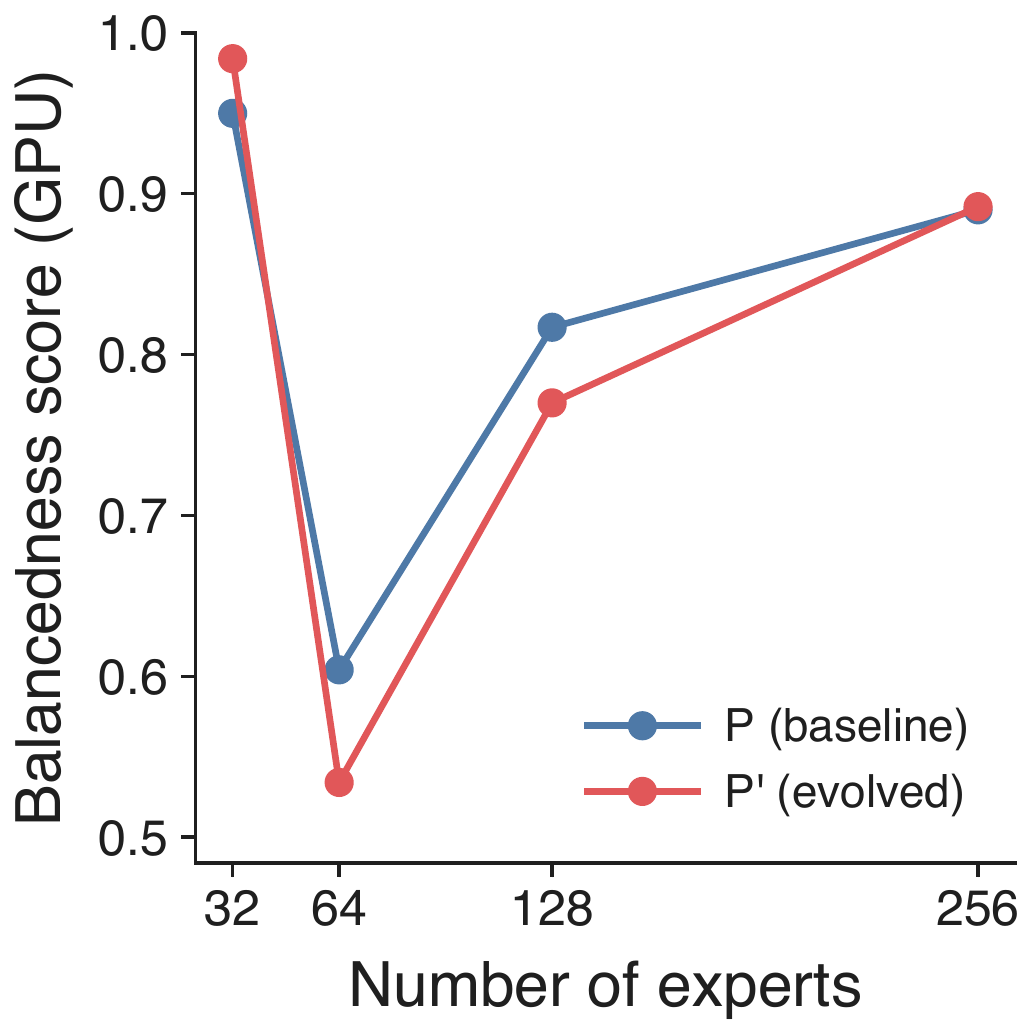}
        \caption{Score w. increasing experts}
        \label{fig:eval:eplb_expert}
    \end{subfigure}
    \caption{On new workloads in EPLB, Opus/AdaEvolve generated $P'$ shows worse balancing score than $P$.}
    \label{fig:eval:eplb_optimality}
    \vspace{-5mm}
\end{figure}
\para{Optimality weaknesses in EPLB.}
The same program also has a  optimality weakness. The  step-by-step replica assignment with a batched shortcut is error-prone  because the load distribution changes after each assignment (lines~6--13). The next best expert may no longer be the one selected in the fixed \texttt{topk} list. Conversely, an expert with much higher load than the others may need several consecutive replicas, but the fixed list can assign too few. The program  returns a valid expert mapping, but the mapping can balance load worse than the initial program.   Figure~\ref{fig:eval:eplb_optimality} shows  this becomes worse  as the number of experts and layers grows. With more experts, small differences in replica choice matter more because there are more near-ties and more high-load experts. With more layers, the same batching error can repeat across layers and accumulate in the final score. In practice, this can under-replicate heavily loaded experts and over-replicate moderately loaded ones,  creating  more load imbalance than the original heuristic.

\para{Scalability weaknesses in TXN.}
TXN scheduling chooses an execution order for transactions. Each transaction reads or writes shared resources, so a poor order can create more lock conflicts and a longer makespan. 
Figure~\ref{fig:txn-code-scalab} shows   the initial program using  a bounded greedy sampler: at each schedule position, it samples a small fixed set of remaining transactions, estimates the cost of adding each one, and commits the best candidate. The AI-evolved program uses   a much larger search procedure. It builds cost tables, scores transaction orderings, generates multiple seeds, refines schedules, runs beam search and bounded A* search, applies large-neighborhood search, and tries hierarchical scheduling. This  helps on small  workloads, but it also turns a bounded greedy policy into a multi-stage search over many possible transaction orders. 
The scalability problem comes from repeatedly analyzing  many partial schedules. The AdaEvolve generated program caches explored prefixes as \texttt{tuple(seq)}, keeps competing prefixes in a priority queue, and copies schedules during each expansion. Its local-refinement loop can also run for many attempts because \texttt{insert\_trials} limits only consecutive failures. As a result, execution time and memory grow with the number of explored prefixes unlike the original program that only keeps one partial schedule and a fixed sampling budget.


\begin{figure}[t]
\centering
\small

\begin{codepanel}{origbg}{Original: one active prefix, bounded sampling per position}
for i in range(0, workload.num_txns - 1):
    ...
    for j in range(0, num_samples):
        ...
        test_seq = txn_seq.copy()
        test_seq.append(t)
        cost = workload.get_opt_seq_cost(test_seq)
        if cost < min_cost:
            min_cost = cost
            min_txn = t
        if done:
            break
    txn_seq.append(min_txn)
\end{codepanel}
\begin{codepanel}{engbg}{GPT/AdaEvolve evolved: unbounded search on partial schedules}
#(1)Memory risk: cache every explored partial schedule.
eval_cache = {}
def eval_cost(seq):
    k = tuple(seq)
    ...
    eval_cache[k] = workload.get_opt_seq_cost(seq)
#(2)Time & memory risk: A* branches over many prefixes.
while pq and expanded < max_nodes:
    ...
    ranked = sorted(rem, key=lambda j: h_cost(seq, rem, j, 0.35))
    for t in ranked[:top_k]:
        ns = seq + [t]
        nr = [x for x in rem if x != t]
        g2 = eval_cost(ns)
        heapq.heappush(pq, (..., tie_score(ns, nr), ..., ns, nr, g2))
#(3)Time risk: trial budget restarts after each improvement.
while tries < insert_trials:
    ...
    if c < best_c:
        ...
        improved_here = True
        break
    tries = 0 if improved_here else tries + 1
\end{codepanel}
    \vspace{-3mm}
\caption{\textbf{TXN-scheduling scalability weakness}}
\label{fig:txn-code-scalab}
    \vspace{-1mm}
\end{figure}
\begin{table}[t]
  \centering
  \resizebox{\columnwidth}{!}{%
  \begin{tabular}{l l l p{2.9cm}}
    \toprule
    \textbf{App} & \textbf{Original$(P)$} & \textbf{AI-evolved$(P')$} & \textbf{Symbol Legend} \\
    \midrule

    TXN
      & $\mathcal{O}(N)$
      & $\mathcal{O}(N^{2})$
      & $N$: \# transactions \\

      EPLB
  & $\mathcal{O}(L E B)$
  & $\mathcal{O}(L E B + L R)$ 
& $L$: \# layers; $E$: \# logical experts; $R$: \# physical experts; 
  $B$: max replicas\\

    \addlinespace
    Prism
      & $\mathcal{O}(M + G)$
      & $\mathcal{O}(M G)$
      & $M$: \# models; $G$: \# GPUs\\

    \bottomrule
  \end{tabular}%
  }
  \caption{Big-$O$ annotation of original program $P$ vs.\ Opus/AdaEvolve evolved program $P'$.}
  \label{tab:scalab-mem-complexity}
      \vspace{-2mm}

\end{table}

\para{Scalability weakness impact.} In production systems, such  scalability  weaknesses can turn the scheduler itself into a bottleneck. A scheduler that spends seconds searching for a slightly better order can delay short transactions that should finish in milliseconds. Under high contention, the evolved scheduler may expand many similar prefixes before returning a schedule, increasing tail latency when the system is already under pressure. Its growing prefix cache and priority queue can also compete with the database engine for memory and risk garbage-collection pressure or out-of-memory failures. Thus, an optimization meant to reduce lock conflicts can instead burn CPU and memory before transactions even begin executing.

We also compare memory complexity across applications using AdaEvolve (Opus-4.6) as $P'$ in Table~\ref{tab:scalab-mem-complexity}. A common pattern is that $P'$ improves evaluator scores by storing more intermediate state: search prefixes in TXN, packing and device-transfer copies in EPLB, and a dense model--GPU cost matrix in Prism. These structures fit small evaluator workloads but grow quickly on larger valid inputs.

\begin{tcolorbox}[colback=green!6!white,colframe=green!50!black,sharp corners=south,boxrule=0.8pt,enhanced,width=\linewidth,arc=3pt,left=4pt,right=4pt,top=4pt,bottom=4pt]
\textbf{\textit{Takeaway 3:}} 
\textit{AI-evolved program trades the lightweight greedy design of the baseline program for more global optimization logic.
This can improve the benchmark objective, but it does not necessarily preserve robustness properties that are not encoded in the evaluator.
Under adversarial workload shifts, the selected evolved programs may expose crashes, hidden peak-memory regressions, execution-time blowups, or lower solution quality than the human-designed baseline.
}

\end{tcolorbox}

\subsection{Mitigation with \sysname}
\label{sec:eval:mitigation}




\begin{table}[t]
\centering
\small
\resizebox{\columnwidth}{!}{%
\begin{tabular}{lccccc}
\toprule
\textbf{}
& \textbf{Cloudcast} $\downarrow$
& \textbf{EPLB}
& \textbf{Prism}
& \textbf{LLM-SQL}
& \textbf{TXN} \\
\midrule

\multicolumn{6}{l}{\textit{Human baseline}} \\
Score
& 626.2
& 0.126
& 21.89
& 0.692
& 2,725 \\

\midrule
\multicolumn{6}{l}{\textit{AdaEvolve (reproduced)}} \\
\rowcolor{green!10}
Score
& 627.1
& 0.141
& 26.25
& 0.730
& 4,081 \\

\midrule
\multicolumn{6}{l}{\textit{AdaEvolve+Prompt fixing}} \\
\rowcolor{green!10}
Score
& 627.1
& 0.137
& 24.22
& 0.734
& 3,185 \\

Correctness
& \ding{55}
& \checkmark
& \checkmark
& \ding{55}
& \checkmark \\

Scal\_Time
& \checkmark
& \ding{55}
& \ding{55}
& \checkmark
& \ding{55} \\

Scal\_Mem
& \checkmark
& \checkmark
& \checkmark
& \checkmark
& \ding{55} \\

Optimality
& \checkmark
& \checkmark
& \ding{55}
& \checkmark
& \checkmark \\

\midrule
\multicolumn{6}{l}{\textit{AdaEvolve+\sysname}} \\
\rowcolor{green!10}
Score
& 627.1
& 0.153
& 21.89
& 0.712
& 2,832 \\

Correctness
& \checkmark
& \checkmark
& \checkmark
& \checkmark
& \checkmark \\

Scal\_Time
& \checkmark
& \checkmark
& \checkmark
& \checkmark
& \checkmark \\

Scal\_Mem
& \checkmark
& \checkmark
& \checkmark
& \checkmark
& \checkmark \\

Optimality
& \checkmark
& \checkmark
& \checkmark
& \checkmark
& \checkmark \\

\bottomrule
\end{tabular}
}
\caption{After mitigation attempt, scores of AI-evolved program and weakness-finding results for Opus/AdaEvolve. 
(\ding{55}~=~weakness found, \checkmark~=~no weakness found).}
\label{tab:mitigation}
\vspace{-2mm}
\end{table}


Based on the weaknesses found by \sysname, we next test whether we can mitigate such weaknesses. We evaluate this in AdaEvolve with Opus-4.6. 

As a baseline we consider \emph{prompt-based patching}.  We modify the original evolution prompt to warn the model about the four weakness types studied in this paper: correctness, scalability-time, scalability-memory, and optimality regressions. The prompt asks the model to avoid changes that may introduce these failures. Appendix~\ref{prompt:mitigation} gives the full prompt. This tests whether telling the model about the weaknesses is enough to prevent them.

We also design a  new strategy, \emph{AdaEvolve+\sysname}, using  \sysname as an explicit checking step inside the evolution loop. The original AdaEvolve loop ranks each candidate program only by its evaluator score. 
In this evolution loop, after each candidate $P'$ is generated, we run \sysname for 100 weakness-finding iterations to search for valid workloads where $P'$ regresses against $P$. If \sysname finds any adversarial workload, we assign the candidate a penalty of $-100$ and return the workload and weakness type as feedback to the model.
In effect, \sysname is part of the CI/CD pipeline for \adrs: a candidate must not only improve the  score, but also pass the weakness checks.


Table~\ref{tab:mitigation} shows that prompt patching is not enough: the generated programs still contain all weakness types across applications. In contrast, with \sysname integrated into AdaEvolve++, the final selected programs no longer expose weaknesses found by \sysname. This robustness, however, reduces the AI-evolved benchmark gains. The original AdaEvolve improves the evaluator score by 19\% on \textsc{Prism} and 49\% on \textsc{TXN-Sched}. With AdaEvolve++, \textsc{Prism} falls back to the human baseline, giving 0\% improvement, and \textsc{TXN-Sched} drops to 3.9\%.

\begin{tcolorbox}[colback=green!6!white,colframe=green!50!black,sharp corners=south,boxrule=0.8pt,enhanced,width=\linewidth,arc=3pt,left=4pt,right=4pt,top=4pt,bottom=4pt]
\textbf{\textit{Takeaway 4:}} 
\textit{Adding \sysname to the AI-evolution loop can filter out candidates with weaknesses, but it also has a tradeoff: once those candidates are penalized, the measured improvement often shrinks, and in some applications disappears. This suggests that to unlock the full potential of AI-evolved system, it requires strong and robust benchmark workloads for evaluation feedback.}
\end{tcolorbox}



\section{Related Work}
\para{AI-evolved solutions for systems.}
Recent work has explored using AI agents to design and optimize system algorithms. \adrs~\cite{Barbarian-2} uses OpenEvolve~\cite{openevolve-original} to iteratively generate, evaluate, and refine code for system applications. AdaEvolve~\cite{adaevolve} improves this loop by adaptively reallocating search effort toward more promising evolutionary directions. Evox~\cite{liu2026evox} further evolves both candidate solutions and the search strategies that produce them. Glia~\cite{glia} proposes an autonomous architecture for designing and optimizing system components such as routing, scheduling, and autoscaling algorithms. Engram~\cite{engram} focuses on improving long-horizon LLM search by reducing local-search bias and preserving useful context across iterations. Self-Defining Systems~\cite{sds-ratul} takes a broader view, using multi-agent systems with long-term memory to operate and improve complex applications over time.
These systems show that AI agents can produce high-scoring system code compared to original human-designed program. 
\sysname takes a different stance, to learn whether these AI-evolved programs remain safe and robust beyond the evaluator workloads used. 
\sysname tests the outputs of AI agents and searches for valid adversarial workloads that expose crashes, resource blowups, or KPI regressions relative to the original program.

\para{Traditional fuzzing.}
Greybox fuzzers use lightweight execution feedback to guide mutation: AFL~\cite{afl} retains inputs that increase edge coverage, AFL++~\cite{afl++} integrates practical AFL-family improvements, and LibAFL~\cite{fioraldi2022libafl} modularizes these components for reuse. Performance-oriented fuzzers replace or augment coverage with resource signals: SlowFuzz~\cite{petsios2017slowfuzz} searches for worst-case complexity inputs, PerfFuzz~\cite{lemieux2018perffuzz} uses multi-dimensional feedback to expose diverse hotspots, and SPIDER~\cite{spider_leo} targets stateful performance bugs in ONOS via dependency-aware modular fuzzing. Grammar-based fuzzers instead preserve input structure: NAUTILUS~\cite{nautilus} and Gramatron~\cite{gramatron} guide grammar-level generation and mutation, FANDANGO~\cite{zamudio2025fandango} evolves grammar-valid inputs under semantic constraints, and GraphFuzz~\cite{green2022graphfuzz} mutates lifetime-aware dataflow graphs for library APIs. 
Unlike traditional greybox fuzzers, \sysname targets AI-evolved system algorithms and searches valid workload spaces with differential oracles to expose regressions between the initial program $P$ and evolved program $P'$.


\section{Discussion and Limitations}


\para{AI-evolved system programs may be harder to audit.}
Our results suggest that AI evolution can improve evaluator scores by replacing simple heuristics with much more complex code, sometimes an order of magnitude longer. This complexity adds hidden assumptions, intermediate state, and interactions between heuristics. Even when the KPI improves, developers may struggle to understand what its resource bounds are, or which workload changes can break it. For system algorithms, simplicity is often part of robustness: a less aggressive human-designed heuristic may preserve operational margins that the evaluator does not measure.

\para{Toward safer AI evolution.} A safer evolution loop could treat each AI-generated change like a CI/CD patch: small, reviewable, and tested before it is accepted. Instead of rewriting a whole algorithm at once, the framework could limit each iteration to a local change, such as one condition or helper function, then run adversarial workloads to check correctness, resource use, and solution quality. This would trade exploration speed for auditability. Improvements may arrive more slowly, but they would be easier to explain, test, and roll back. Our mitigation experiment provides initial evidence for this direction.

\para{Imperfect root-cause analysis.} \sysname confirms weaknesses through program execution, but still uses an agent to explain root causes. These explanations are useful but not always consistent. The agent may only describe a nearby symptom instead of identify the actual root cause. Future work could strengthen this step with cross-checking, dynamic slicing, or delta debugging to better ground each root-cause report in execution evidence.

\begin{table*}[t]
\centering
\caption{Representative changes made by AI-evolved TXN schedulers. AdaEvolve replaces the original greedy sampler with a global continuous optimizer, while Engram keeps the greedy structure but adds heavier sampling, restarts, and local search.}
\label{tab:txn_code_compare}
\setlength{\tabcolsep}{2pt}
\renewcommand{\arraystretch}{0.95}
\begin{tabular}{@{}p{0.32\textwidth}p{0.32\textwidth}p{0.32\textwidth}@{}}
\toprule
\multicolumn{1}{c}{\textbf{Original}} &
\multicolumn{1}{c}{\textbf{AdaEvolve}} &
\multicolumn{1}{c}{\textbf{Engram}} \\
\midrule
\begin{lstlisting}[style=pycompact]
def get_best_schedule(workload, num_seqs):
    def greedy_sample(num_samples, sample_rate):
        # random starting transaction
        start = random.randint(
            0, workload.num_txns - 1)
        txn_seq = [start]
        remaining = list(range(workload.num_txns))
        remaining.remove(start)
        for _ in range(workload.num_txns - 1):
            min_cost, min_txn = 100000, -1
            holdout = []
            # sample next-transaction choices
            for _ in range(num_samples):
                t = random.choice(remaining)
                holdout.append(t)
                remaining.remove(t)
                cost = workload.get_opt_seq_cost(
                    txn_seq + [t])
                if cost < min_cost:
                    min_cost, min_txn = cost, t
                if len(remaining) == 0:
                    break
            # greedily append best sampled txn
            txn_seq.append(min_txn)
            holdout.remove(min_txn)
            remaining.extend(holdout)
        return workload.get_opt_seq_cost(txn_seq), txn_seq
    return greedy_sample(10, 1.0)
\end{lstlisting}
&
\begin{lstlisting}[style=pycompact]
def get_best_schedule(workload, num_seqs):
    (*@\diff{from scipy.optimize import dual\_annealing}@*)
    (*@\diff{import numpy as np}@*)
    n = workload.num_txns
    (*@\diff{def priorities\_to\_schedule(x):}@*)
        # continuous priorities -> permutation
        return list(np.argsort(x))
    (*@\diff{def objective(x):}@*)
        schedule = priorities_to_schedule(x)
        return workload.get_opt_seq_cost(schedule)
    (*@\diff{bounds = [(0, n) for \_ in range(n)]}@*)
    best_cost = float('inf')
    best_schedule = None
    (*@\diff{\# multiple annealing restarts}@*)
    for restart in range(max(2, num_seqs // 5)):
        if restart == 0:
            x0 = greedy_initial_by_write_count()
        else:
            x0 = np.random.uniform(0, n, n)
        (*@\diff{result = dual\_annealing(}@*)
            objective,
            bounds=bounds,
            x0=x0,
            maxiter=300,
            no_local_search=False,
            seed=restart)
        schedule = priorities_to_schedule(result.x)
        cost = workload.get_opt_seq_cost(schedule)
        if cost < best_cost:
            best_cost, best_schedule = cost, schedule
    (*@\diff{\# final adjacent-swap refinement}@*)
    for i in range(min(10, len(best_schedule)-1)):
        test = best_schedule[:]
        test[i], test[i+1] = test[i+1], test[i]
        cost = workload.get_opt_seq_cost(test)
        if cost < best_cost:
            best_cost, best_schedule = cost, test
    return best_cost, best_schedule
\end{lstlisting}
&
\begin{lstlisting}[style=pycompact]
def get_best_schedule(workload, num_seqs):
    def greedy_sample(num_samples, sample_rate,
                      seed_start=None):
        (*@\diff{\# heuristic or seeded start}@*)
        if seed_start is None:
            start = txn_with_most_writes(workload)
        else:
            start = seed_start
        txn_seq = [start]
        remaining = list(range(workload.num_txns))
        remaining.remove(start)
        for _ in range(workload.num_txns - 1):
            min_cost, min_txn = 100000, -1
            holdout = []
            for _ in range(num_samples):
                t = random.choice(remaining)
                holdout.append(t)
                remaining.remove(t)
                cost = workload.get_opt_seq_cost(
                    txn_seq + [t])
                (*@\diff{\# deterministic tie-breaking}@*)
                if cost < min_cost:
                    min_cost, min_txn = cost, t
                elif cost == min_cost and \
                     writes(t) > writes(min_txn):
                    min_txn = t
            txn_seq.append(min_txn)
            holdout.remove(min_txn)
            remaining.extend(holdout)
        return workload.get_opt_seq_cost(txn_seq), txn_seq
    (*@\diff{\# heavier greedy sampling + restarts}@*)
    candidates = []
    candidates.append(greedy_sample(40, 1.0))
    for _ in range(6):
        s = random.randint(0, workload.num_txns - 1)
        candidates.append(greedy_sample(40, 1.0, s))
    candidates.sort(key=lambda x: x[0])
    (*@\diff{\# local search on best candidates}@*)
    best_cost, best_seq = float('inf'), None
    for cost, seq in candidates[:3]:
        cost, seq = local_search_2opt(seq, 40)
        cost, seq = local_search_oropt_simple(seq, 20)
        if cost < best_cost:
            best_cost, best_seq = cost, seq
    return best_cost, best_seq
\end{lstlisting}
\\
\bottomrule
\end{tabular}
\end{table*}

\section{Conclusions}
AI-driven system evolution can substantially help systems, but these gains may come with hidden weaknesses. This paper argues that evaluating AI-evolved system programs requires testing not only average-case performance on fixed benchmarks, but also worst-case behavior across valid workload spaces. We present \sysname, a weakness searching framework that compares an AI-evolved program against its human baseline. 
Results show that \sysname exposes distinct hidden weaknesses across evolved programs and applications. We propose that automated weakness discovery is a necessary step toward safely deploying AI-generated systems code.

\newpage
\bibliographystyle{acm}
\bibliography{adrs_risk}

@article{adaevolve,
  title={AdaEvolve: Adaptive LLM driven zeroth-order optimization},
  author={Cemri, Mert and Agrawal, Shubham and Gupta, Akshat and Liu, Shu and Cheng, Audrey and Mang, Qiuyang and Naren, Ashwin and Erdogan, Lutfi Eren and Sen, Koushik and Zaharia, Matei and others},
  journal={arXiv preprint arXiv:2602.20133},
  year={2026}
}

@article{barbarians,
  title={Barbarians at the gate: How ai is upending systems research},
  author={Cheng, Audrey and Liu, Shu and Pan, Melissa and Li, Zhifei and Wang, Bowen and Krentsel, Alex and Xia, Tian and Cemri, Mert and Park, Jongseok and Yang, Shuo and others},
  journal={arXiv preprint arXiv:2510.06189},
  year={2025}
}

@misc{eplb,
  title        = {{Expert Parallelism Load Balancer (EPLB)}},
  author       = {{DeepSeek AI}},
  year         = {2024},
  howpublished = {\url{https://github.com/deepseek-ai/eplb}},
}

@misc{openevolve-original,
  title        = {{OpenEvolve}},
  author       = {{Sharma, Asankhaya}},
  year         = {2025},
  howpublished = {\url{https://github.com/algorithmicsuperintelligence/openevolve}},
}

@misc{basf-alphaevolve,
  title        = {{How BASF Manages Thousands of Supply Chain Decisions with AlphaEvolve's Agentic Algorithms}},
  author       = {Priese, Benjamin and Nawalgaria, Anant},
  year         = {2026},
  howpublished = {\url{https://cloud.google.com/blog/products/ai-machine-learning/how-basf-manages-thousands-of-supply-chain-decisions-with-alphaevolve}},
}

@article{txn_scheduling,
  title={Towards optimal transaction scheduling},
  author={Cheng, Audrey and Kabcenell, Aaron and Chan, Jason and Shi, Xiao and Bailis, Peter and Crooks, Natacha and Stoica, Ion},
  journal={Proceedings of the VLDB Endowment},
  volume={17},
  number={11},
  pages={2694--2707},
  year={2024},
  publisher={VLDB Endowment}
}

@inproceedings{cloudcast,
  title={Cloudcast:$\{$High-Throughput$\}$,$\{$Cost-Aware$\}$ overlay multicast in the cloud},
  author={Wooders, Sarah and Liu, Shu and Jain, Paras and Mo, Xiangxi and Gonzalez, Joseph E and Liu, Vincent and Stoica, Ion},
  booktitle={21st USENIX Symposium on Networked Systems Design and Implementation (NSDI 24)},
  pages={281--296},
  year={2024}
}

@article{llm_sql,
  title={Optimizing llm queries in relational data analytics workloads},
  author={Liu, Shu and Biswal, Asim and Kamsetty, Amog and Cheng, Audrey and Schroeder, Luis G and Patel, Liana and Cao, Shiyi and Mo, Xiangxi and Stoica, Ion and Gonzalez, Joseph E and others},
  journal={Proceedings of Machine Learning and Systems},
  volume={7},
  year={2025}
}

@article{prism,
  title={Prism: Unleashing gpu sharing for cost-efficient multi-llm serving},
  author={Yu, Shan and Xing, Jiarong and Qiao, Yifan and Ma, Mingyuan and Li, Yangmin and Wang, Yang and Yang, Shuo and Xie, Zhiqiang and Cao, Shiyi and Bao, Ke and others},
  journal={arXiv preprint arXiv:2505.04021},
  year={2025}
}

@inproceedings{afl++,
  title={$\{$AFL++$\}$: Combining incremental steps of fuzzing research},
  author={Fioraldi, Andrea and Maier, Dominik and Ei{\ss}feldt, Heiko and Heuse, Marc},
  booktitle={14th USENIX workshop on offensive technologies (WOOT 20)},
  year={2020}
}

@inproceedings{property_fuzz,
  title={QuickCheck: a lightweight tool for random testing of Haskell programs},
  author={Claessen, Koen and Hughes, John},
  booktitle={Proceedings of the fifth ACM SIGPLAN international conference on Functional programming},
  pages={268--279},
  year={2000}
}

@article{alphaevolve,
  title={Alphaevolve: A coding agent for scientific and algorithmic discovery},
  author={Novikov, Alexander and V{\~u}, Ng{\^a}n and Eisenberger, Marvin and Dupont, Emilien and Huang, Po-Sen and Wagner, Adam Zsolt and Shirobokov, Sergey and Kozlovskii, Borislav and Ruiz, Francisco JR and Mehrabian, Abbas and others},
  journal={arXiv preprint arXiv:2506.13131},
  year={2025}
}

@inproceedings{pagedattention,
  title={Efficient memory management for large language model serving with pagedattention},
  author={Kwon, Woosuk and Li, Zhuohan and Zhuang, Siyuan and Sheng, Ying and Zheng, Lianmin and Yu, Cody Hao and Gonzalez, Joseph and Zhang, Hao and Stoica, Ion},
  booktitle={Proceedings of the 29th symposium on operating systems principles},
  pages={611--626},
  year={2023}
}

@article{lost-in-the-middle,
  author       = {Nelson F. Liu and
                  Kevin Lin and
                  John Hewitt and
                  Ashwin Paranjape and
                  Michele Bevilacqua and
                  Fabio Petroni and
                  Percy Liang},
  title        = {Lost in the Middle: How Language Models Use Long Contexts},
  journal      = {CoRR},
  volume       = {abs/2307.03172},
  year         = {2023}
}

@article{adrs-db,
  title={AI-Driven Research for Databases},
  author={Cheng, Audrey and Ng, Harald and Kabcenell, Aaron and Bailis, Peter and Zaharia, Matei and Ma, Lin and Shi, Xiao and Stoica, Ion},
  journal={arXiv preprint arXiv:2604.06566},
  year={2026}
}

@article{sds-ratul,
  title={Self-Defining Systems},
  author={Anderson, Thomas and Mahajan, Ratul and Peter, Simon and Zettlemoyer, Luke},
  year={2025}
}

@misc{skydiscover,
    title={SkyDiscover: A Flexible Framework for AI-Driven Scientific and Algorithmic Discovery},
    author={GitHub},
    howpublished={\url{https://github.com/skydiscover-ai/skydiscover}},
year={2026}
}

@misc{cocoevolve,
    title={CoCoEvolve: What If a Coding Agent Could Optimize Your AI Systems?},
    author={Snowflake},
    howpublished={\url{https://www.snowflake.com/en/blog/engineering/optimize-snowflake-ai-systems-cocoevolve/}},
year={2026}
}

@inproceedings{wang2020not,
  title={Not All Coverage Measurements Are Equal: Fuzzing by Coverage Accounting for Input Prioritization.},
  author={Wang, Yanhao and Jia, Xiangkun and Liu, Yuwei and Zeng, Kyle and Bao, Tiffany and Wu, Dinghao and Su, Purui},
  booktitle={NDSS},
  year={2020}
}

@misc{opus-4.6,
    title={Introducing Claude Opus 4.6},
    author={Anthropic},
    howpublished={\url{https://www.anthropic.com/news/claude-opus-4-6}},
year={2026}
}

@misc{gpt-5,
    title={Introducing GPT‑5},
    author={OpenAI},
    howpublished={\url{https://openai.com/index/introducing-gpt-5/}},
year={2025}
}

@misc{hypothesis,
    title={Hypothesis: a property-based testing library for Python},
    author={Hypothesis},
    howpublished={\url{https://github.com/HypothesisWorks/hypothesis/}},
year={2025}
}

@article{engram,
  title={Improving Coherence and Persistence in Agentic AI for System Optimization},
  author={Karimi, Pantea and Noorbakhsh, Kimia and Alizadeh, Mohammad and Balakrishnan, Hari},
  journal={arXiv preprint arXiv:2603.21321},
  year={2026}
}

@article{glia,
  title={Glia: A human-inspired ai for automated systems design and optimization},
  author={Hamadanian, Pouya and Karimi, Pantea and Nasr-Esfahany, Arash and Noorbakhsh, Kimia and Chandler, Joseph and ParandehGheibi, Ali and Alizadeh, Mohammad and Balakrishnan, Hari},
  journal={arXiv preprint arXiv:2510.27176},
  year={2025}
}

@inproceedings{spider_leo,
  title={SPIDER: Fuzzing for Stateful Performance Issues in the ONOS Software-Defined Network Controller},
  author={Li, Ao and Padhye, Rohan and Sekar, Vyas},
  booktitle={2025 IEEE Conference on Software Testing, Verification and Validation (ICST)},
  pages={1--12},
  year={2025},
  organization={IEEE}
}

@inproceedings{petsios2017slowfuzz,
  title={Slowfuzz: Automated domain-independent detection of algorithmic complexity vulnerabilities},
  author={Petsios, Theofilos and Zhao, Jason and Keromytis, Angelos D and Jana, Suman},
  booktitle={Proceedings of the 2017 ACM SIGSAC conference on computer and communications security},
  pages={2155--2168},
  year={2017}
}

@inproceedings{lemieux2018perffuzz,
  title={Perffuzz: Automatically generating pathological inputs},
  author={Lemieux, Caroline and Padhye, Rohan and Sen, Koushik and Song, Dawn},
  booktitle={Proceedings of the 27th ACM SIGSOFT international symposium on software testing and analysis},
  pages={254--265},
  year={2018}
}

@inproceedings{gramatron,
  title={Gramatron: Effective grammar-aware fuzzing},
  author={Srivastava, Prashast and Payer, Mathias},
  booktitle={Proceedings of the 30th acm sigsoft international symposium on software testing and analysis},
  pages={244--256},
  year={2021}
}

@inproceedings{nautilus,
  title={NAUTILUS: Fishing for deep bugs with grammars.},
  author={Aschermann, Cornelius and Frassetto, Tommaso and Holz, Thorsten and Jauernig, Patrick and Sadeghi, Ahmad-Reza and Teuchert, Daniel},
  booktitle={NDSS},
  volume={19},
  pages={337},
  year={2019}
}

@article{zamudio2025fandango,
  title={FANDANGO: evolving language-based testing},
  author={Zamudio Amaya, Jos{\'e} Antonio and Smytzek, Marius and Zeller, Andreas},
  journal={Proceedings of the ACM on Software Engineering},
  volume={2},
  number={ISSTA},
  pages={894--916},
  year={2025},
  publisher={ACM New York, NY, USA}
}

@inproceedings{green2022graphfuzz,
  title={Graphfuzz: Library api fuzzing with lifetime-aware dataflow graphs},
  author={Green, Harrison and Avgerinos, Thanassis},
  booktitle={Proceedings of the 44th International Conference on Software Engineering},
  pages={1070--1081},
  year={2022}
}

@inproceedings{fioraldi2022libafl,
  title={Libafl: A framework to build modular and reusable fuzzers},
  author={Fioraldi, Andrea and Maier, Dominik Christian and Zhang, Dongjia and Balzarotti, Davide},
  booktitle={Proceedings of the 2022 ACM SIGSAC Conference on Computer and Communications Security},
  pages={1051--1065},
  year={2022}
}

@article{afl,
  title={American fuzzy lop-whitepaper},
  author={Zalewski, Micha{\l}},
  journal={Retrieved September},
  volume={1},
  pages={2022},
  year={2016}
}

@article{Barbarian-2,
  title={Let the Barbarians In: How AI Can Accelerate Systems Performance Research},
  author={Cheng, Audrey and Liu, Shu and Pan, Melissa and Li, Zhifei and Agarwal, Shubham and Cemri, Mert and Wang, Bowen and Krentsel, Alexander and Xia, Tian and Park, Jongseok and others},
  journal={arXiv preprint arXiv:2512.14806},
  year={2025}
}

@article{liu2026evox,
  title={Evox: Meta-evolution for automated discovery},
  author={Liu, Shu and Agarwal, Shubham and Maheswaran, Monishwaran and Cemri, Mert and Li, Zhifei and Mang, Qiuyang and Naren, Ashwin and Boneh, Ethan and Cheng, Audrey and Pan, Melissa Z and others},
  journal={arXiv preprint arXiv:2602.23413},
  year={2026}
}

@article{map-elites,
  title={Illuminating search spaces by mapping elites},
  author={Mouret, Jean-Baptiste and Clune, Jeff},
  journal={arXiv preprint arXiv:1504.04909},
  year={2015}
}

@article{random_fuzz,
  title={An empirical study of the reliability of UNIX utilities},
  author={Miller, Barton P and Fredriksen, Lars and So, Bryan},
  journal={Communications of the ACM},
  volume={33},
  number={12},
  pages={32--44},
  year={1990},
  publisher={ACM New York, NY, USA}
}

@inproceedings{Domagoj19-fudge,
author = {Babi\'{c}, Domagoj and Bucur, Stefan and Chen, Yaohui and Ivan\v{c}i\'{c}, Franjo and King, Tim and Kusano, Markus and Lemieux, Caroline and Szekeres, L\'{a}szl\'{o} and Wang, Wei},
title = {FUDGE: fuzz driver generation at scale},
year = {2019},
isbn = {9781450355728},
publisher = {Association for Computing Machinery},
address = {New York, NY, USA},
url = {https://doi.org/10.1145/3338906.3340456},
doi = {10.1145/3338906.3340456},
booktitle = {Proceedings of the 2019 27th ACM Joint Meeting on European Software Engineering Conference and Symposium on the Foundations of Software Engineering},
pages = {975–985},
numpages = {11},
location = {Tallinn, Estonia},
series = {ESEC/FSE 2019}
}

@inproceedings{cadar08-klee,
author = {Cadar, Cristian and Dunbar, Daniel and Engler, Dawson},
title = {KLEE: unassisted and automatic generation of high-coverage tests for complex systems programs},
year = {2008},
publisher = {USENIX Association},
address = {USA},
booktitle = {Proceedings of the 8th USENIX Conference on Operating Systems Design and Implementation},
pages = {209–224},
numpages = {16},
location = {San Diego, California},
series = {OSDI'08}
}

@article{Cadar08-exe,
author = {Cadar, Cristian and Ganesh, Vijay and Pawlowski, Peter M. and Dill, David L. and Engler, Dawson R.},
title = {EXE: Automatically Generating Inputs of Death},
year = {2008},
issue_date = {December 2008},
publisher = {Association for Computing Machinery},
address = {New York, NY, USA},
volume = {12},
number = {2},
issn = {1094-9224},
url = {https://doi.org/10.1145/1455518.1455522},
doi = {10.1145/1455518.1455522},
journal = {ACM Trans. Inf. Syst. Secur.},
month = dec,
articleno = {10},
numpages = {38}
}

@inproceedings{Nguyen22-bedivfuzz,
author = {Nguyen, Hoang Lam and Grunske, Lars},
title = {BeDivFuzz: integrating behavioral diversity into generator-based fuzzing},
year = {2022},
isbn = {9781450392211},
publisher = {Association for Computing Machinery},
address = {New York, NY, USA},
url = {https://doi.org/10.1145/3510003.3510182},
doi = {10.1145/3510003.3510182},
booktitle = {Proceedings of the 44th International Conference on Software Engineering},
pages = {249–261},
numpages = {13},
location = {Pittsburgh, Pennsylvania},
series = {ICSE '22}
}
\clearpage
\onecolumn
\section{Appendix}
\para{\sysname Prompt Details}
\begin{PromptBlock}[prompt:param_infer]{Workload Space Inference}
You are analyzing an \adrs application to infer the full input space for adversarial bug discovery.

\vspace{4pt}
\pcode{run\_workload.py} is a fixed per-app harness.
\begin{itemize}[leftmargin=*, itemsep=2pt]
    \item This file is fixed and \pkey{cannot be changed}.
    \item It defines the workload dictionary schema.
    \item Parameter names in \pcode{grammar\_workload} must exactly match the keys read by \pcode{workload.get("key", ...)}.
\end{itemize}

\pimp{Your task:} Produce a JSON object with the following fields.

\begin{itemize}[leftmargin=*, itemsep=2pt]
    \item \pcode{grammar\_config}: structural parameters that define the deployment environment, such as number of GPUs, replicas, or nodes.
    \item \pcode{grammar\_workload}: per-invocation algorithm inputs, such as request distributions, transaction sequences, query data, or load tensors.
    \item \pcode{constraints}: cross-parameter constraints written as plain English strings.
    \item \pcode{notes}: any other information needed to generate valid inputs.
\end{itemize}

\pimp{Important:}
\begin{itemize}[leftmargin=*, itemsep=2pt]
    \item Structural parameters may be hardcoded constants in the original program, but for risk discovery we want to \pkey{explore different configurations}.
    \item Give each non-fixed structural parameter a valid range to try.
    \item Do \pkey{not} mark a parameter as fixed unless the evaluator is hard-wired to that value.
    \item Example ranges: \pcode{num\_gpus in [8, 16, 32, 64]}, \pcode{num\_nodes in [1, 2, 4, 8]}.
\end{itemize}

\pimp{Evaluator-hardwired parameters:}
\begin{itemize}[leftmargin=*, itemsep=2pt]
    \item If a module-level constant is used directly inside the evaluator, mark it as fixed.
    \item Examples: \pcode{num\_physical\_experts = NUM\_REPLICAS}, \pcode{gpu\_load = total\_physical\_load.view(..., NUM\_GPUS, ...)}.
    \item If a parameter is only passed into the algorithm, assign a realistic range to explore.
\end{itemize}

\pimp{For each parameter include:}
\begin{itemize}[leftmargin=*, itemsep=2pt]
    \item \pcode{name}
    \item \pcode{type}: \pcode{int}, \pcode{float}, \pcode{str}, \pcode{list}, or \pcode{tensor}
    \item \pcode{range}: human-readable range
    \item Use \pcode{values} only for true categorical parameters, such as topology type.
\end{itemize}

\pimp{\# CRITICAL:} Parameter \pcode{name} fields in \pcode{grammar\_workload} must exactly match the keys that \pcode{run\_workload.py} reads through \pcode{workload.get("key")} calls. Wrong names silently fall through to defaults and kill exploration diversity.

\vspace{4pt}
\pimp{\# OUTPUT FORMAT:} Return \pkey{only} a JSON block followed by a Python code block. No other text.
\end{PromptBlock}

\begin{PromptBlock}[prompt:root_cause]{Explain Root Cause of Weaknesses}

You are analyzing a bug in an AI-optimized program \pcode{P'}.

\plabel{Anomalous execution profile:}
\begin{verbatim}
{{anomalous_lines_summary}}
\end{verbatim}

\vspace{4pt}
\pimp{Your task:} In exactly \pkey{2--3 sentences}, state the root cause of this bug.

\pimp{Your explanation must include:}
\begin{itemize}[leftmargin=*, itemsep=2pt]
    \item Which specific branch or code path in \pcode{P'} is triggered by this input.
    \item Why that path produces incorrect or inefficient behavior.
    \item What property of \pcode{(c, w)} triggers it.
\end{itemize}

\vspace{4pt}
\pimp{\# OUTPUT FORMAT:} Return \pkey{only plain text}. No code, no headers, and no bullet points.
\end{PromptBlock}

\para{Naive Agent Baseline}

\begin{PromptBlock}[prompt:naive_agent]{Single Prompt to Find All Weakness Types}
You are helping find bugs in an AI-optimized program \pcode{P'} by generating new valid \pcode{(c, w)} input pairs. Each generated input will be compared against the baseline program \pcode{P} on the same workload.

\vspace{4pt}
\pimp{Bug types to target:}
\begin{itemize}[leftmargin=*, itemsep=2pt]
    \item \pkey{Correctness}: \pcode{P'} crashes, raises an exception, or times out while \pcode{P} succeeds.
    \item \pkey{Scalability-time}: \pcode{P'} is significantly slower than \pcode{P} on the same input; a timeout in \pcode{P'} always counts.
    \item \pkey{Scalability-memory}: \pcode{P'} uses significantly more peak memory than \pcode{P}, measured by \pcode{tracemalloc}.
    \item \pkey{Optimality}: \pcode{P} produces better output quality than \pcode{P'} by more than 5\% on any quality metric.
\end{itemize}

\vspace{4pt}
\pimp{Target-specific search guidance:}
\begin{itemize}[leftmargin=*, itemsep=2pt]
    \item \pkey{Correctness}: try boundary values, empty lists, zero, negative values, maximum values, unusual parameter combinations, missing keys, \pcode{None}, division-by-zero cases, and index-out-of-bounds cases.
    \item \pkey{Scalability-time}: push size/count parameters toward their maximums, and try inputs that expose worse algorithmic complexity, expensive fallback paths, long chains, maximum fan-out, or adversarial orderings.
    \item \pkey{Scalability-memory}: try wide inputs, high-cardinality categorical values, large tensors, large batches, deep object graphs, and inputs that cause \pcode{P'} to materialize data that \pcode{P} streams or avoids.
    \item \pkey{Optimality}: try skewed distributions, adversarial datasets, all-identical values, zero-valued fields, extreme ratios, and inputs where \pcode{P'} may trade output quality for speed.
\end{itemize}
\vspace{4pt}
\pimp{Your task:} Generate exactly new \pcode{(c, w)} pairs by mutating the seed.
\end{PromptBlock}

\para{Mitigation Prompts}
\begin{PromptBlock}[prompt:mitigation]{Prompt Patching Mitigation (Cloudcast application as an example)}
You are an expert in cloud infrastructure optimization. Your task is to evolve the \pcode{search\_algorithm(src, dsts, G, num\_partitions)} function to minimize overall data transfer cost across multiple clouds.

\vspace{4pt}
\pimp{Optimization goal:}
\begin{itemize}[leftmargin=*, itemsep=2pt]
    \item Minimize total data transfer cost across multiple cloud networks.
    \item Efficiently broadcast input data from \pcode{src} to multiple destination nodes \pcode{dsts}.
    \item Use parallel paths and overlapping transfers when they reduce cost.
    \item Use the \pcode{BroadCastTopology} class and \pcode{make\_nx\_graph} function to identify low-cost routes.
\end{itemize}

\pimp{Search guidance:}
\begin{itemize}[leftmargin=*, itemsep=2pt]
    \item Be diverse and innovative, as long as the generated program optimizes the given metric.
    \item Reduce redundant transfers.
    \item Balance load across networks.
    \item Exploit multi-network topologies to reduce broadcast cost.
    \item Prefer strategies that remain reliable on unseen network topologies.
\end{itemize}

\end{PromptBlock}

\end{document}